\pdfoutput=1

\documentclass[letterpaper, 10 pt, conference]{ieeeconf} \IEEEoverridecommandlockouts
\let\labelindent\relax

\usepackage{graphics}
\usepackage{cite}
\usepackage{amsmath,amssymb,amsfonts}
\usepackage{graphicx}
\usepackage{textcomp}
\usepackage{enumitem}
\usepackage{xcolor}
\usepackage{tikz}
\usepackage{verbatim}
\usetikzlibrary{shapes.gates.logic.US,trees,positioning,arrows}
\usepackage{todonotes}
\usepackage{algorithmicx}
\usepackage[ruled]{algorithm}
\usepackage{algpseudocode}

\usepackage{multirow}
\usepackage{mathabx}

\usepackage{caption}
\usepackage{subcaption}
\DeclareCaptionLabelSeparator{periodspace}{.\quad}
\graphicspath{{./images/}}

\def\BibTeX{{\rm B\kern-.05em{\sc i\kern-.025em b}\kern-.08em
    T\kern-.1667em\lower.7ex\hbox{E}\kern-.125emX}}
\begin{document}

\title{\huge Compact Belief State Representation for Task Planning
\thanks{
$^1$The authors are with Istituto Italiano di Tecnologia, Genoa, Italy. \newline
$^2$Evgenii Safronov is also with Department of Informatics, Bioengineering, Robotics and Systems Engineering, Universit\`a di Genova, Genova, Italy
\texttt{evgenii.safronov@iit.it}
}
}

\author{{Evgenii Safronov$^{1,2}$, Michele Colledanchise$^1$, and Lorenzo Natale$^1$}}

%

\newcommand{\mnt}{\emptyset}
\newcommand{\msR}{\mathbb{R}}
\newcommand{\msS}{\mathbb{S}}
\newcommand{\msF}{\mathbb{F}}

\newcommand{\nand}{\underline{AND}}
\newcommand{\nor}{\underline{OR}}
\newcommand{\nlit}{\underline{LIT}}

\newcommand{\lE}{$\mathbb{E}~$}
\newcommand{\lI}{$\mathbb{I}~$}
\newcommand{\lM}{$\mathbb{M}~$}

\newcommand{\lEn}{$\mathbb{E}$}
\newcommand{\lIn}{$\mathbb{I}$}
\newcommand{\lMn}{$\mathbb{M}$}

\newcommand{\nt}{$\emptyset$}
\newcommand{\sR}{$\mathbb{R}$}
\newcommand{\sS}{$\mathbb{S}$}
\newcommand{\sF}{$\mathbb{F}$}

\maketitle

\begin{abstract}
Task planning in a probabilistic belief space generates complex and robust execution policies in domains affected by state uncertainty. The performance of a task planner relies on the belief space representation of the world. However, such representation becomes easily intractable as the number of variables and execution time grow. To address this problem, we developed a novel belief space representation based on the Cartesian product and union operations over belief substates. These two operations and single variable assignment nodes form And-Or directed acyclic graph of Belief States (AOBSs). We show how to apply actions with probabilistic outcomes and how to measure the probability of conditions holding true over belief states. We evaluated AOBSs performance in simulated forward state space exploration. We compared the size of AOBSs with the size of Binary Decision Diagrams (BDDs) that were previously used to represent belief state. We show that AOBSs representation more compact than a full belief state and it scales better than BDDs for most of the cases.
\end{abstract}

\section{Introduction}
	With the advances in perception, navigation, and manipulation tasks, robots are becoming capable to \emph{act} in uncertain environments. However, to \emph{plan} in such environments, the robots need an additional deliberation capability that combines the tasks above in a given policy. Task planning aims at creating such policies from a set of actions, conditions, and other constraints. Modern research addresses the task planning in partially observable and uncertain environments\cite{kaelbling2013integrated,colledanchise2020act} formulating the problem in the \emph{belief space}, a probabilistic distribution over physical states of the system. In this paper, we limit ourselves to discrete distributions with a finite number of physical states. Belief state represents a collection of physical states and corresponding non-zero probabilities. Belief substate is a subset of this collection. Basic operations in the task planning include: applying an action over belief substate and inference in a form of evaluating a condition over belief state. Forward search represents the simplest approach in task planning. One of its limitations is the exponential growth of the state size with the search depth. Tackling this problem could extend a horizon of long-term task planning and improve performance in existing tasks.
    The main contribution of this work is a novel belief state representation based on an And-Or directed acyclic graph which we call And Or Belief State (AOBS). AOBS exactly describes a discrete probabilistic belief state, having a much smaller size than the collection of physical states. Task planning routines such as acting on a belief state, selecting a substate by a boolean conditionary function can be applied directly on AOBS keeping the compressed form.
\section{Related Works}
\label{s:rel-works}
		A Probabilistic Belief State (PBS) could be treated as real-valued function over set of discrete variables $0 \leq p(s) \leq 1, p(s) \in \mathbb{R}$, where $s$ is a state vector. Sometimes in the literature the term \emph{belief state} refers to a collection of physical states without known probabilities. In this case, the boolean function $b(s)$ describes a belief state. Compressed representations of boolean functions over discrete arguments are well presented in the literature. A standard approach for that is a Binary Decision Diagram (BDD) \cite{akers1978binary}. Reduced Ordered Binary Decision Diagrams (OBDD) are a restricted form of BDD\cite{10.1109/TC.1986.1676819}. In this paper, we consider OBDD only.

BDDs proven themselves as a promising data structure in symbolic planning\cite{speck2018symbolic}, graphical models\cite{mateescu2008and}, bayesian networks\cite{DAL2017411}, and stochastic constraint programming\cite{babaki2019compiling}. However, their performance is sensitive to correct variable order\cite{10.1109/TC.1986.1676819}. While better variables ordering could be guessed by a human designer in other applications, in task planning it becomes impossible to automate. Moreover, it depends on the statespace exploration result, which is the goal of task planning routine.

    \begin{figure}[t]
    \centering
\includegraphics[width=0.7\columnwidth]{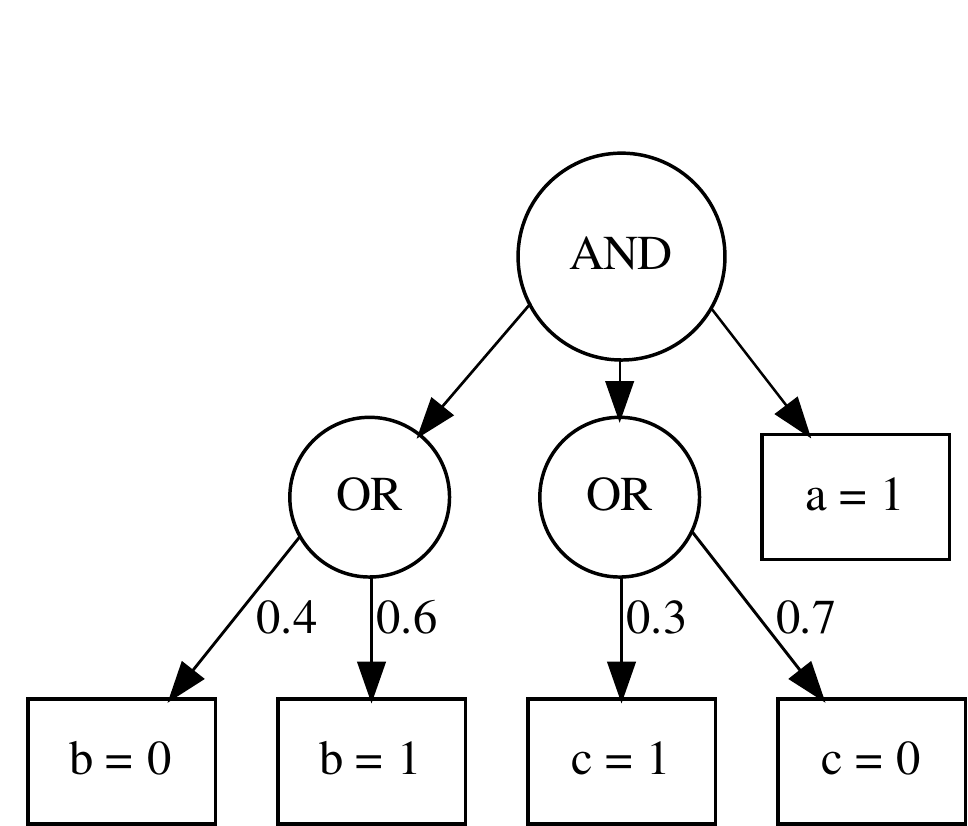}

         \caption{Example of AOBS graph for belief state defined in Table \ref{tbl:simple-example}. \nand{} node makes Cartesian product of its children substates, while \nor{} node makes a union.}
         \label{fig:simple-example}
         \vspace{-1.8em}
    \end{figure}
    BDDs were suggested as a compact representation of the collection of physical states\cite{Bertoli2001HeuristicS}. However, sometimes BDDs could not capture the probability distribution over physical states. Take a look at a simple example of probabilistic belief state (Table. \ref{tbl:simple-example}, Fig. \ref{fig:simple-example}).
    \begin{table}[h]
    \centering
        \caption{Example of probabilistic belief state with 4 non-zero probability $P$ physical states over 3 variables $a$, $b$, $c$.}
    \begin{subtable}[t]{.48\columnwidth}
        \centering
    \caption{Full belief state tabular definition.}
    \begin{tabular}{| c | c | c | c |}
    \hline
	    P 		& a & b & c \\ \hline
	    0.28		& 0 & 0 & 0 \\ \hline
	    0.42		& 0 & 1 & 0 \\ \hline
	    0.12		& 0 & 0 & 1 \\ \hline
	    0.18		& 0 & 1 & 1 \\ \hline
    \end{tabular}
        \label{tbl:simple-example}
    \end{subtable}
    ~
    \begin{subtable}[t]{.48\columnwidth}
        \centering
    \caption{An example of belief substate of PBS on the Table \ref{tbl:simple-example}.}
    \begin{tabular}{| c | c | c |}
    \hline
	    P 		 & b & c \\ \hline
	    0.28		 & 0 & 0 \\ \hline
	    0.42		 & 1 & 0 \\ \hline
    \end{tabular}
        \label{tbl:simple-example-substate}
    \vspace{-1em}
    \end{subtable}

    \end{table}
BDD representing this belief space without probabilities consists of only one node $a = 1$ according to the reduction rules. It is not enough to distinguish between different physical states, hence we could not represent probabilistic belief space without excluding some of the reduction rules. Various modifications of BDD were developed. Edge-valued Decision Diagrams\cite{vrudhula1996edge} allow value mechanism in a form of function factorized over edges. Unfortunately, they could fail in case different physical states have the same probability. Multiterminal Decision Diagrams could distinguish between all physical states in case we add a terminal for each physical state. However, it would result in $O(N)$ memory consumption, where $N$ is a number of physical states in the system. And Or Multivalued Decision Diagrams\cite{mateescu2008and} benefits in performance over BDDs when there is a predefined pseudo tree of problem statespace decomposition. This is not exactly the case of belief state where each physical states contain all the variables (it depends on every variable). Nevertheless, the authors noticed the impossibility to reduce the And-Or graph to Decision Diagram in some cases of \emph{weighted graphical models}.

\section{And Or Belief State}
	The main source of belief state expansion in task planning is acting on state. We can start from a physical state that represents current robot sensoric input and generate belief state by applying actions, e.g., from Belief Behavior Tree\cite{safronov2020iros}. A probabilistic action could be represented by the union of its outcomes. It can be written down in a tabular form, take a look at example on Table \ref{tbl:action-pbs}. Each row of the table corresponds to one of probabilistic outcomes, e.g. with probability $0.7$ set $Y=2$, $Z=1$. Let us take a simple example of belief state (Table \ref{tbl:bs-initial}, two physical states) where the rest of statespace (variable $X$) remains the same for all possible values of $Y$, $Z$. If we apply the action (Table \ref{tbl:action-pbs}) on this state, we notice that the result basically overwrites the probabilities and values of $Y$,$Z$ in the initial state. Note, that initial behavior state is a cartesian product of its $X=0$ and $Y,Z$ parts. This is a key finding that allows to build AOBS ``in place'' preserving compact size of representation. We later show how acting operation generalizes for more complicated Cartesian products and substates of a belief state, selected by a condition.
	\begin{table}[h]
    \centering
        \caption{An example of acting on a probabilistic belief state.}
    \begin{subtable}{.3\columnwidth}
        \centering
    \caption{Initial belief state. Each row represents a physical state with its probability}
    \begin{tabular}{| c | c | c | c |}
    \hline
	    P 		& X & Y & Z \\ \hline
	    0.4		& 0 & 0 & 0 \\ \hline
	    0.6		& 0 & 1 & 0 \\ \hline
    \end{tabular}
        \label{tbl:bs-initial}
    \end{subtable}
    ~
    \begin{subtable}{.3\columnwidth}
        \centering
    \caption{An action with probabilistic outcomes. Each row is a single outcome}
    \begin{tabular}{| c | c | c |}
    \hline
	    P 		 & Y & Z \\ \hline
	    0.7		 & 2 & 1 \\ \hline
	    0.3		 & 2 & 0 \\ \hline
    \end{tabular}
        \label{tbl:action-pbs}
    \end{subtable}
    ~
    \begin{subtable}{.3\columnwidth}
        \centering
    \caption{Result of acting on belief state (Table \ref{tbl:bs-initial}) by (Table \ref{tbl:action-pbs})}
    \begin{tabular}{| c | c | c | c |}
    \hline
	    P 		& X & Y & Z \\ \hline
	    0.7		& 0 & 2 & 1 \\ \hline
	    0.3		& 0 & 2 & 0 \\ \hline
    \end{tabular}
        \label{tbl:bs-after}
    \vspace{-1em}
    \end{subtable}
    \end{table}

    We chose the Cartesian product ($\times$) and union ($\cup$) as two primary operations over belief substates to form a tree and then a directed acyclic graph. Each internal node of the graph corresponds to one of these two operations over its children, while leaf nodes are single-variable assignments. While creating such a representation of minimal size from the tabular definition of belief state seems to be a difficult task, we show that action application could be done efficiently. 
    \par In this section we describe the structure of AOBS and most important operations on it. Each physical state is a state variable assignment function $s_i := {v_{j} = u_{ij} | v_{j} \in V}$, $V$ is a set of all variables. There is no limitation for value state space $u_{ij} \in U$, it is not obliged to defined before operating on AOBS. PBS is a discrete probability distribution over physical states: $\bigcup_i \{p_i, s_i\}$, where $p_i$ is a probability of physical state $s_i$.
    In this work, by term \emph{subgraph} $G(N)$ of a node $N$ we mean a part of directed acyclic graph that could be achieved from $N$, similar to term \emph{subtree} of a tree. Each subgraph $G(N)$ of each node $N$ in AOBS corresponds to a substate $S(N)$ of a probabilistic belief state. With substate we indicate not only a subset of all physical states collections, but also projection to some variable subspace. In this work, $\Omega(S) \subseteq V$ denotes the variable subset of some belief substate $S$. Table \ref{tbl:simple-example-substate} contains a definition of belief substate $S_e$ from a belief state on Table \ref{tbl:simple-example},  $\Omega(S_e) = \{b,c\}$. $S_e$ could be factorized e.g., as

    $$\left\lbrace\lbrace0.4, b=0\rbrace \cup \{0.6, b=1\}\right\rbrace \times \{0.7, c=0\}$$
     Union operation $\cup$ over two belief substates merges them as merging two sets of pairs $\{p_i, s_i\}$. Cartesian product $\times$ of two belief substates $S_1$ and $S_2$ is defined as follows:
	\begin{multline}
	S_1 = \{p_i, s_i\}_{ 0 \leq i < N}, ~
	S_2 = \{p_j, s_j\}_{ 0 \leq j < M}, \\
	S_1 \times S_2 = \{p_i \cdot p_j, s_i \times s_j\}
	\end{multline}
    To apply union correctly on $S_1$ and $S_2$, they should belong to exactly same variable subspaces, i.e., $\Omega(S_1) = \Omega(S_2)$, while for Cartesian product on $S_1$ and $S_2$ they should lay in different variable subspaces, i.e., $\Omega(S_1) \cap \Omega(S_2) = \emptyset$.
      Note, that probability factors ($0.4$,~$0.6$,~$0.7$ in the example above) could be assigned to the substates in many ways that after applying all operations we result in a correct PBS.

    \paragraph{Literal nodes} In this paper, we call each single variable assignment a \emph{literal}. Each physical state could be described by exactly $|V|$ literals, one per each variable. \emph{Literal} nodes (\nlit) in AOBS graph have \emph{no children}. They contain information about single variable assignment e.g., $a = 0$. A belief substate of \nlit{} node $N$ is simply $S(N) = \{1, v(N) = u(N)\}$, $v(N)$ and $u(N)$ are variable and value stored in $N$. Basically, \nlit{}{} node represents minimal fraction of belief state. $N \in Lit(G)$ means that node $N$ is \nlit{} node.
    \paragraph{Internal nodes} Internal nodes a of graph are either \nand{} or \nor{} nodes. Similarly, $N \in And(G)$ means that $N$ is an \nand{} node and $N \in Or(G)$ means that $N$ is an \nor{} node.  \nor{} node applies union operations over belief substates of its children, while \nand{} node applies Cartesian product:
	\begin{multline}
	N \in Or(G) \Rightarrow
	 S(N) = \bigcup_{ M_i \in Children(N) }  S(M_i), \\
	 N \in And(G) \Rightarrow S(N) = \bigtimes_{ M_i \in Children(N) } S(M_i)
	\end{multline}

\begin{figure}[t!]
    \centering
    \begin{subfigure}[t]{0.49\columnwidth}
        \centering
\includegraphics[width=0.9\columnwidth]{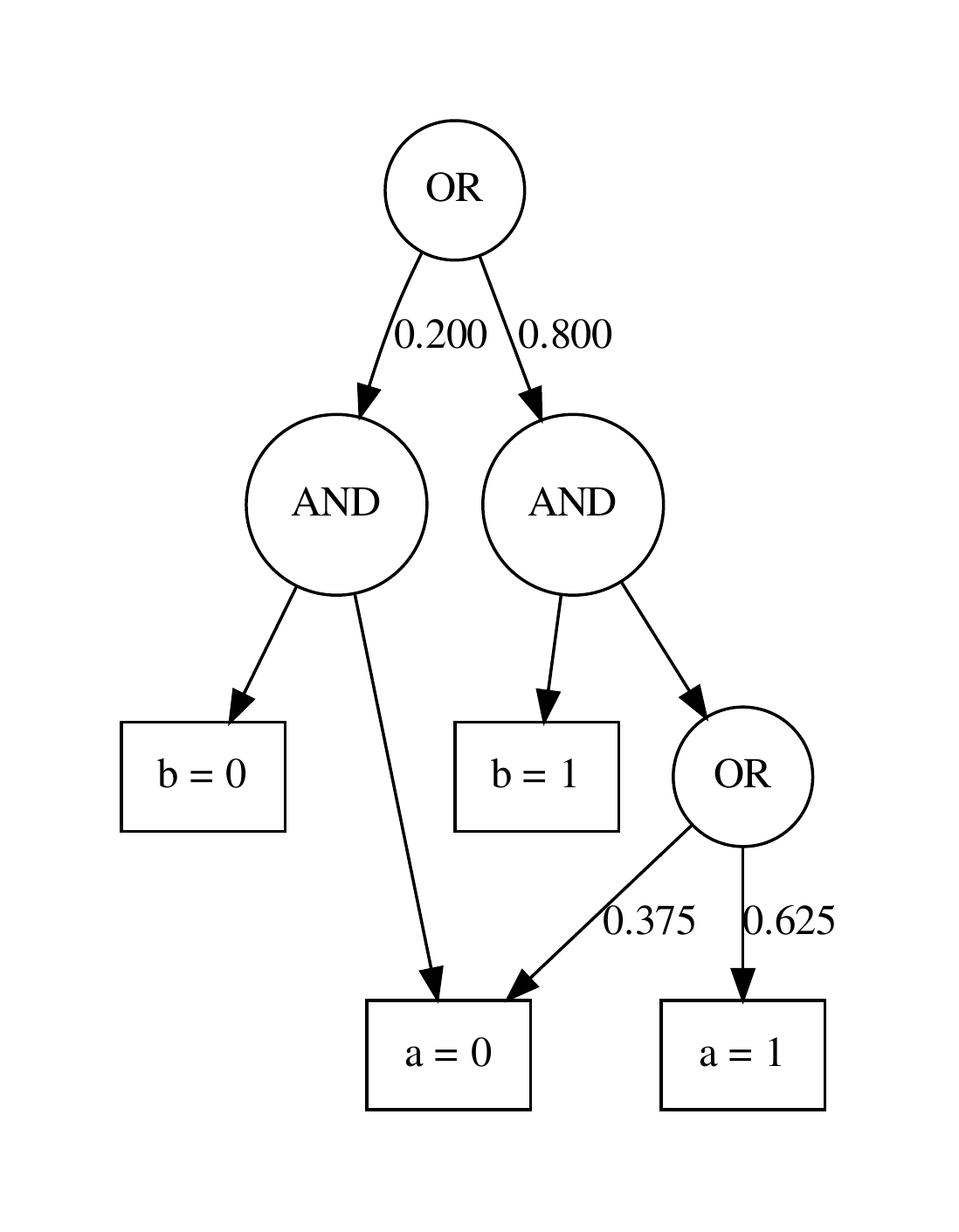}
    \end{subfigure}%
    ~
    \begin{subfigure}[t]{0.49\columnwidth}
\includegraphics[width=0.9\columnwidth]{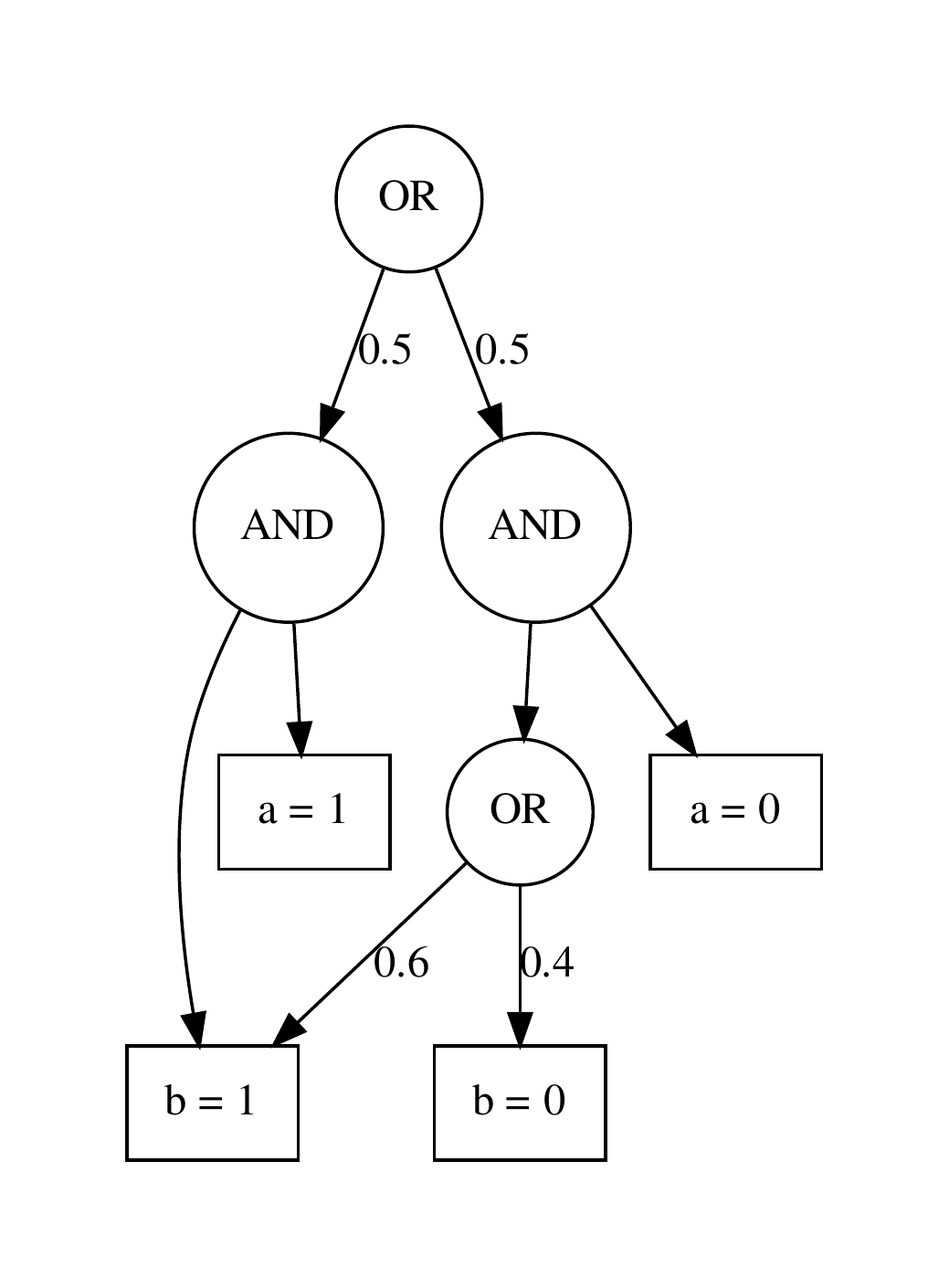}

    \end{subfigure}
    \vspace{-2em}
    \caption{
    Equivalent AOBS representations of belief state from Table \ref{tbl:example-equiv}.
    \vspace{-1.5em}
    \label{fig:example-equiv}
      }
     \end{figure}

 Look at the AOBS example (Figure \ref{fig:simple-example}, Table \ref{tbl:simple-example}). The first \nor{} node defines belief substate $\{0.4, b=0\} \cup \{0.6, b=1\}$. Similarly, the second \nor{} node corresponds to $\{0.3, c=0\} \cup \{0.7, c=1\}$. Finally, \nand{} node makes a Cartesian product
 \begin{multline}
 \{1, a=1\} \times \left\lbrace\{0.4, b=0\} \cup \{0.6, b=1\}\right\rbrace \times \\ \times \left\lbrace\{0.3, c=0\} \cup \{0.7, c=1\}\right\rbrace.
 \end{multline} The result of applying all operations exactly corresponds to the Table \ref{tbl:simple-example}. The recursive procedure of recovering a belief state as a collection of physical states is described in Section \ref{s:conditions}. Unlike BDDs, the AOBS representation could be not unique even for the minimal size of the graph (see an example in Figure \ref{fig:example-equiv}). An algorithm for generating a compact AOBS representation from PBS in the form of a collection of physical states was out of this research scope.
    \begin{table}[h]
        \centering
         \caption{Example of probabilistic belief state with 3 physical states which could be represented by different AOBS graphs with similar sizes.  }
    \begin{tabular}{| c | c | c |}
    \hline
	    P 	& a & b \\ \hline
	    0.2	& 0 & 0 \\ \hline
	    0.3	& 0 & 1 \\ \hline
	    0.5	& 1 & 1 \\ \hline
    \end{tabular}

        \label{tbl:example-equiv}
    \end{table}

    To find out if the belief state described by AOBS contains some physical state $\{v_i = u_i\}$, the condition $\bigwedge_i \{v_i = u_i\}$ must holds true with probability greater than zero. We describe an algorithm for evaluating a condition over belief state later.

    Limiting ourselves in this search for most common operations in task planning, we highlight how the OR operation on two belief states in form of AOBS could be implemented. To implement the logical OR operation (with some probabilities), one needs to add an \nor{} node as root and attach the roots of each input AOBS as children of this new node. With this simple operation, some physical states will be captured twice (if the intersection of input belief states was not empty). We could remove such redundancy with additional operators, such as logical AND operation. However, duplicated physical states do not affect the correctness of inference and other operations described in the paper. Therefore, logical AND operator was out of the scope of this paper. 
\section{Acting on a And Or Belief State}
\label{s:acting}
\begin{algorithm}
        \caption{Recursive labeling procedure}\label{alg:colorize}
        \begin{algorithmic}[1]
            \Function{labeling}{$node$, $labels$, $C$}\\
\Comment{$labels$ argument shares already calculated labels}
\\
\Comment{here \textbf{return} $X$ also sets $labels[node] := X$}
			\If {$node \in labels$}
				\State \textbf{return} $labels[node]$
			\EndIf
			\If {$is\_literal(node)$}
				\If {$var(node) \in \Omega(C)$}
					\If {$C(node)$}
						\State \textbf{return} \lI
					\Else
						\State \textbf{return} \lE
					\EndIf
				\Else
					\State \textbf{return} \lI
				\EndIf
	        \ElsIf {$is\_and(node)$}
	       		\State $has\_mixed := False$
	        		\For{$c \in children(node)$}
	        			\State $labels[c] \gets labeling(c, labels, C)$
	        			\If {$labels[c] = \mathbb{E}$}
	        				\State \textbf{return} \lE
	        			\EndIf
	        			\If {$colors[c] = \mathbb{M}$}
	        				\State $has\_mixed \gets True$
	        			\EndIf
	        		\EndFor
	        		\If{$has\_mixed$}
	        			\State \textbf{return} \lM
	        		\Else
	        			\State \textbf{return} \lI
	        		\EndIf
	        	\Else \Comment{$is\_or(node)$}
	        		\State $has\_inc = False$
	        		\State $has\_exc = False$
	        		\For{$c \in children(node)$}
	        			\State $labels[c] \gets labeling(c, labels, C)$
	        			\If {$labels[c] \in \{\mathbb{E},\mathbb{M}\}$}
	        				\State $has\_exc = True$
	        			\EndIf
	        			\If {$labels[c] \in \{\mathbb{I},\mathbb{M}\}$}
	        				\State $has\_inc \gets True$
	        			\EndIf
	        		\EndFor
	        		\If{$has\_exc$ and $has\_inc$}
	        			\State \textbf{return} \lM
	        		\ElsIf {$has\_inc$}
	        			\State \textbf{return} \lI
	        		\Else
	        			\State \textbf{return} \lE
	        		\EndIf
	        	\EndIf
            \EndFunction
        \end{algorithmic}
    \end{algorithm}
	In this section, we describe an algorithm for acting on an AOBS substate keeping it compact. An action could be applied not only to the whole belief state but to its substate described by some \emph{condition}. A condition is a boolean function of a physical state.  In this work, we limit conditions to a product of boolean functions over single variables:
	\begin{equation}
	C(a,b,c,..) = f_a(a) \wedge f_b(b) \wedge f_c(c) \wedge~ ...
	\label{eq:C}
	\end{equation} This limitation allow us to select literals which \emph{must} be true (e.g. $\{v_i = u_{ij} \| f_{v_i}(u_{ij}) = true \} $). As we discussed before, a subgraph of AOBS corresponds to a belief substate. Informally, we have to find a node in AOBS whose substate contains true literals from condition variables and all the other variables from the action. Then, an action could be performed completely inside this substate (probably, on part of this substate). In general, there could be many such subgraphs. Therefore the main difficulty is to correctly find all such subgraphs in the AOBS.
    \par

\begin{figure*}[t!]
    \centering
    \begin{subfigure}[t]{0.6\columnwidth}
        \centering
\includegraphics[width=0.99\columnwidth]{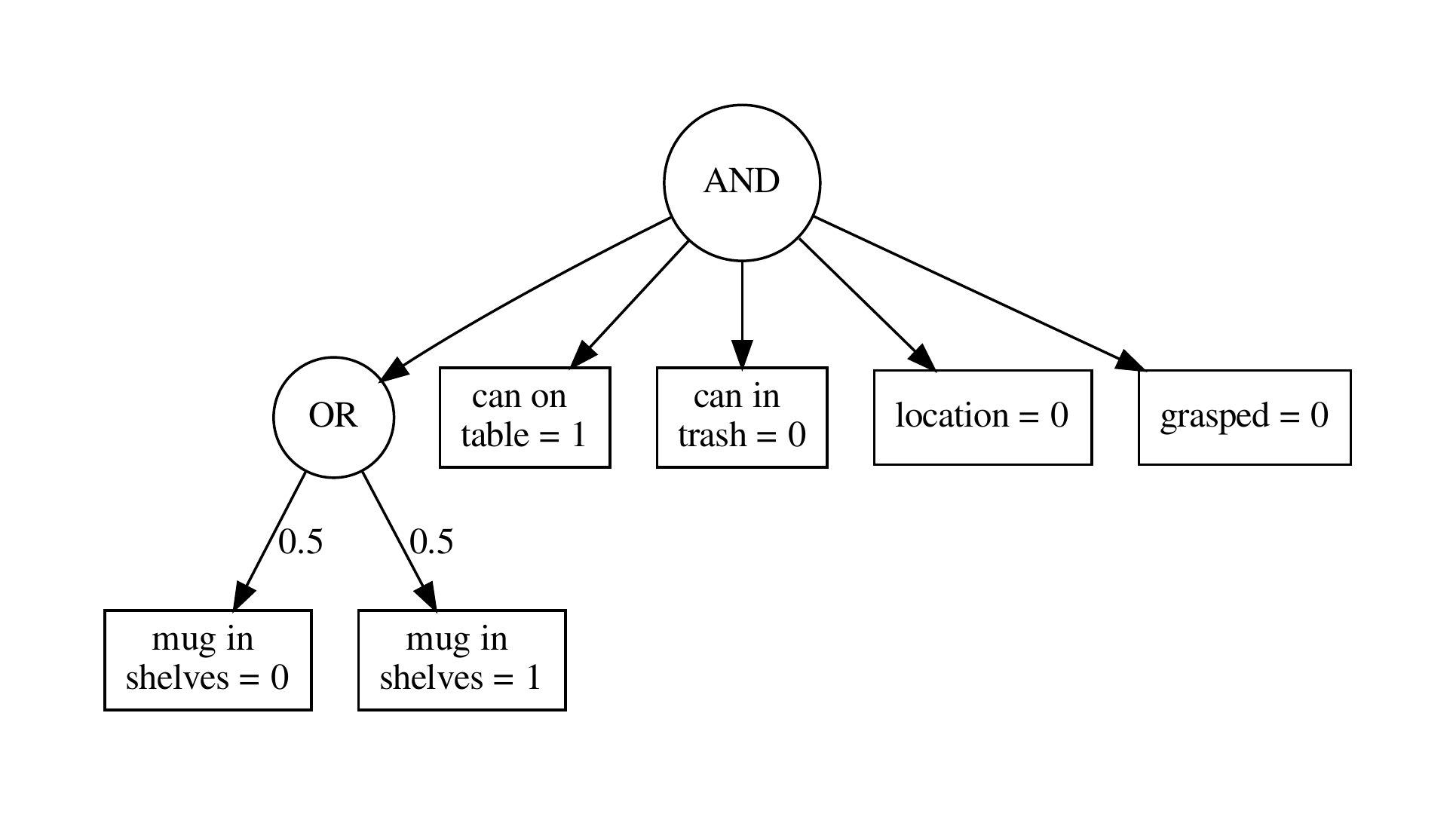}
         \caption{This is an initial belief states as AOBS graph. The only unknown condition is whether object "mug" in shelves or not.}
         \label{fig:aobs1}
    \end{subfigure}%
    ~
    \begin{subfigure}[t]{0.6\columnwidth}
        \centering
\includegraphics[width=0.99\columnwidth]{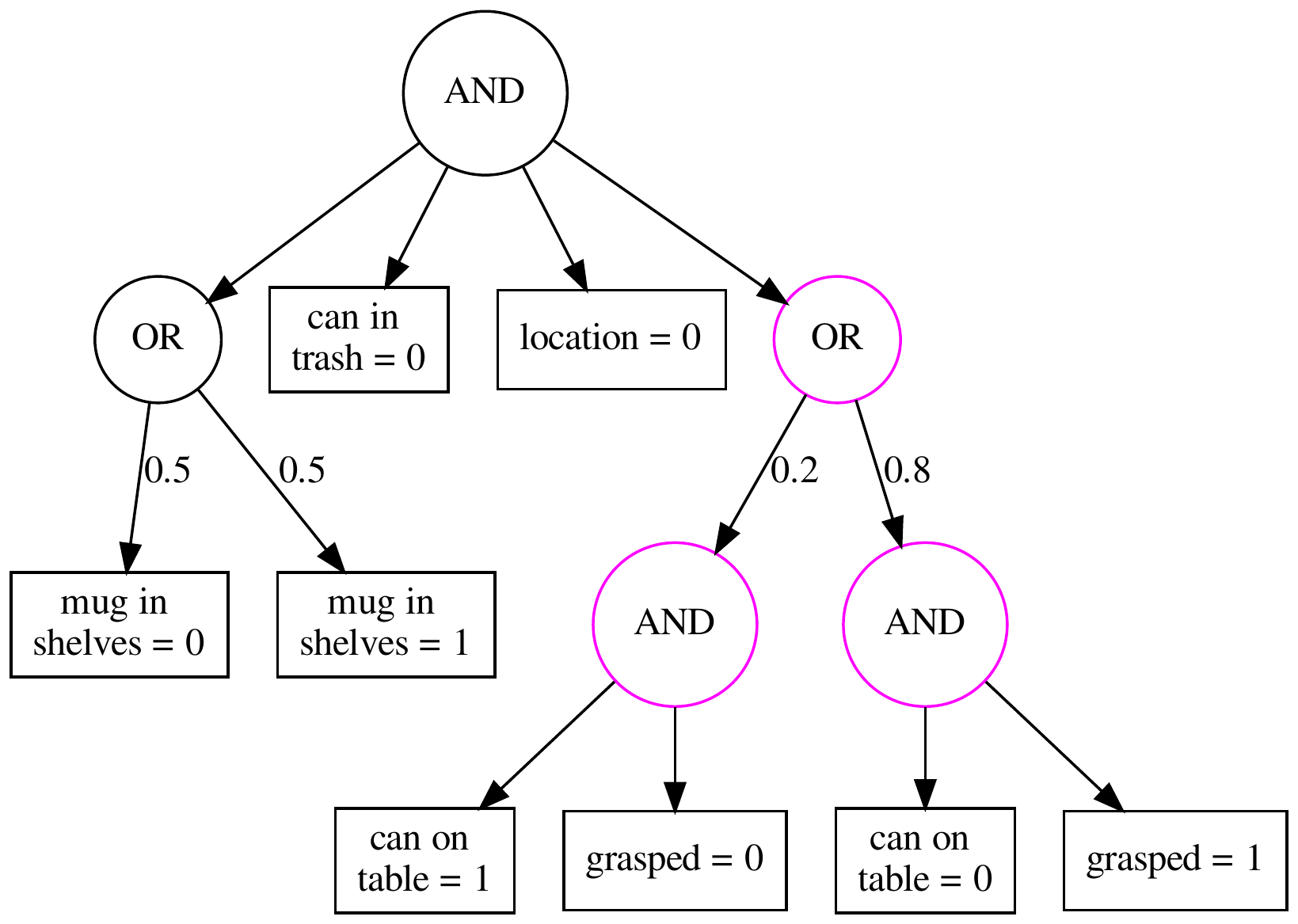}
         \caption{This describes result of applying "grasp the can" action to the whole belief state. Part of belief state which contains action result is painted in purple. Grasping action has two possible results. One, "can on table" is false and "grasped" is true illustrate successful grasping with probability 0.8. The other is "can on table" is true and nothing "grasped" with probability 0.2, illustrates failure of action.}
         \label{fig:aobs2}
    \end{subfigure}%
    ~
    \begin{subfigure}[t]{0.6\columnwidth}
        \centering
\includegraphics[width=0.99\columnwidth]{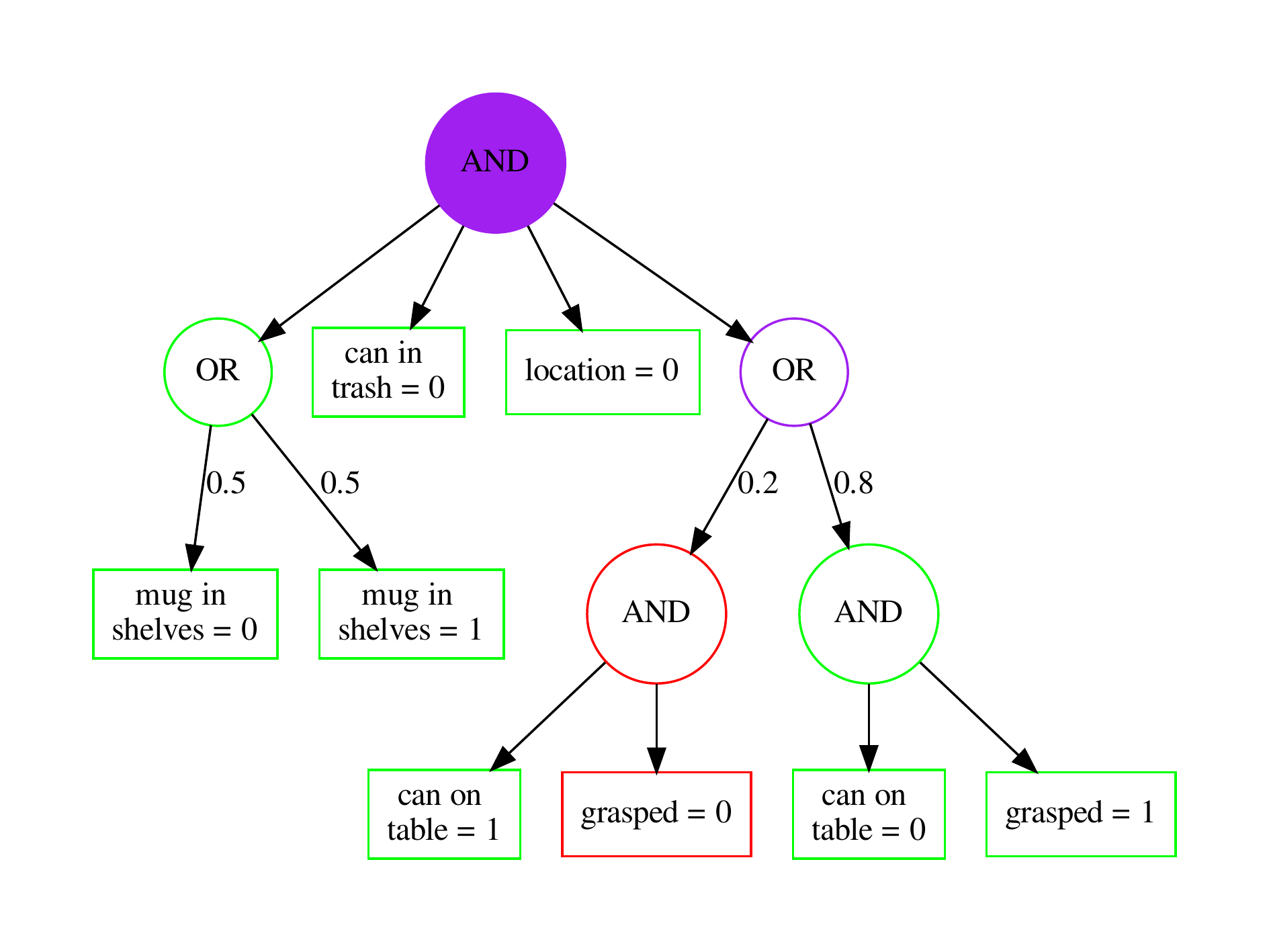}
         \caption{Then we want to put the can into the trash, but the precondition for it is to have a grasped can ("grasped = 1"). In this case, we have a part of belief state where we should apply action and the other part we should not. Red color corresponds to label \lE, green to \lI, and purple to \lM. Minimal subcluster node is filled with color. }
         \label{fig:aobs3}
    \end{subfigure}%
    \\
    \begin{subfigure}[t]{0.6\columnwidth}
\includegraphics[width=0.99\columnwidth]{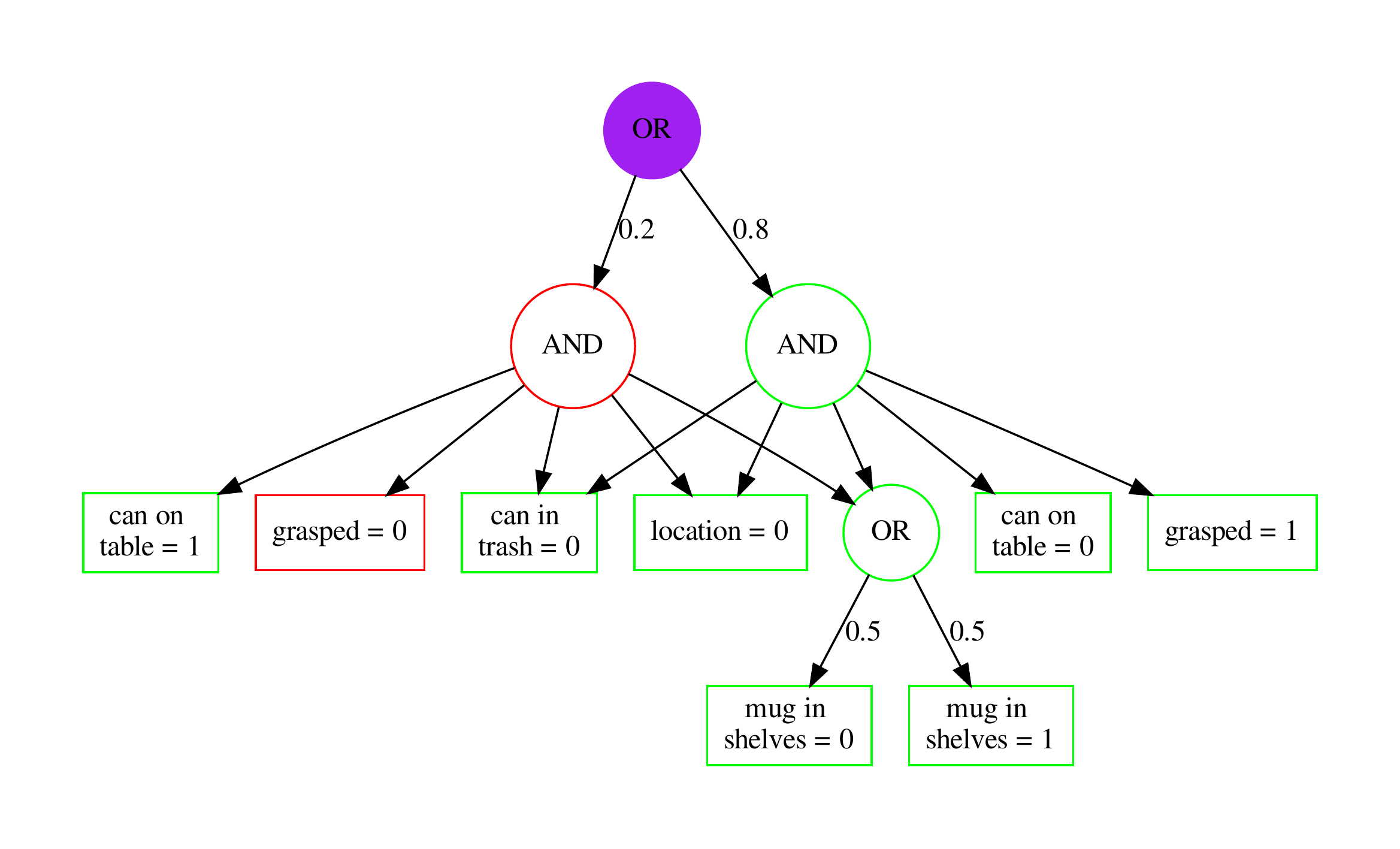}
   \caption{Due to the fact that our minimal subcluster is mixed and has mixed children, we had to perform isolation subroutine. Now we have isolated sub graph, highest green \nand{}, which corresponds exactly to the belief substate where we should apply an action.}
            \label{fig:aobs4}
    \end{subfigure}%
      ~
    \begin{subfigure}[t]{0.6\columnwidth}
        \centering
\includegraphics[width=0.99\columnwidth]{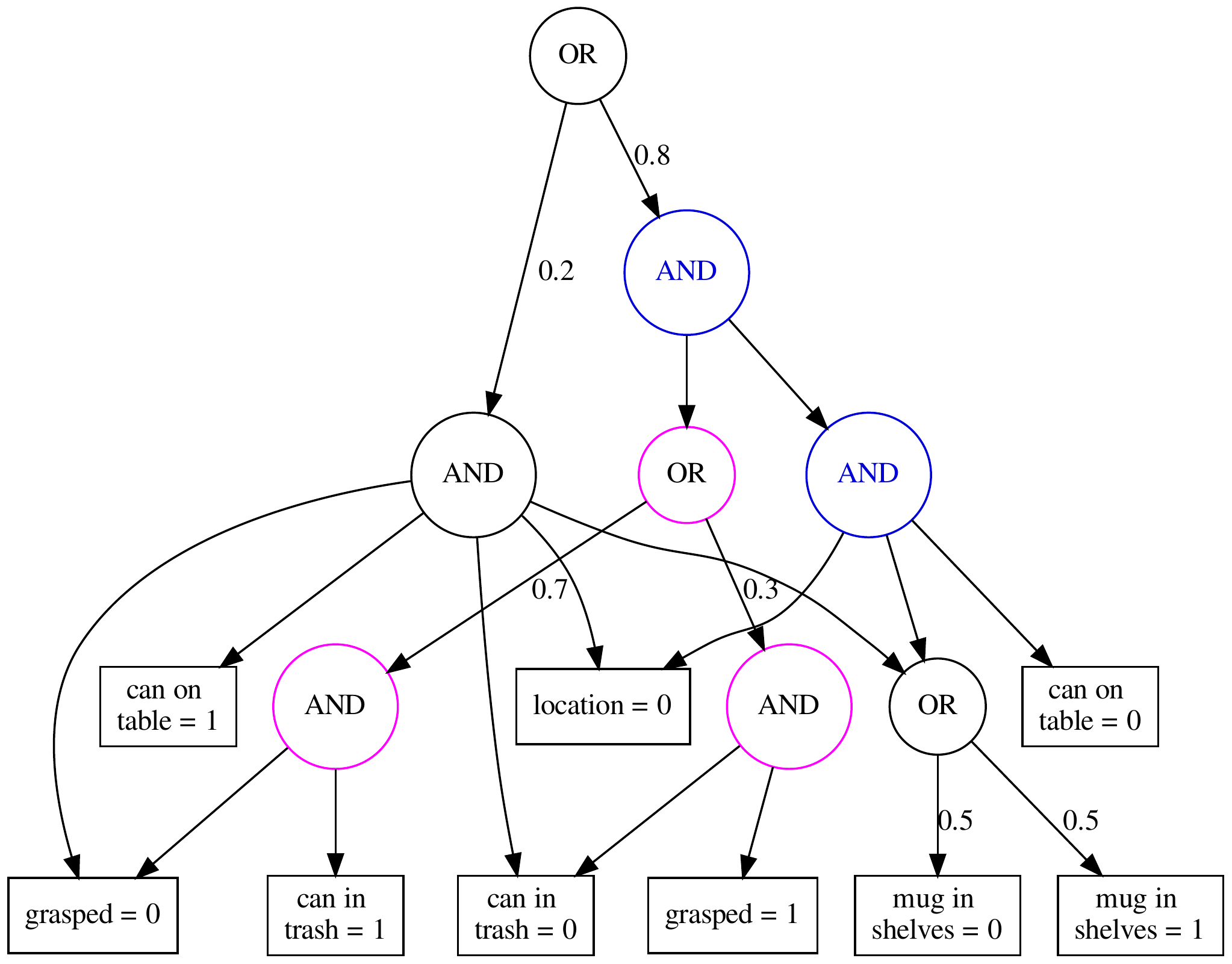}
         \caption{We acted on a belief substate by replacing "grasped" and "can in trash" part from green isolated branch and inserting \nand{} node with action result. Action result subgraph is marked with light purple color.  Now we need to apply normalization procedure, as we have  two \nand{} nodes (marked with blue), one is the child of the other.}
         \label{fig:aobs5}
    \end{subfigure}%
    ~
    \begin{subfigure}[t]{0.6\columnwidth}
        \centering
\includegraphics[width=0.99\columnwidth]{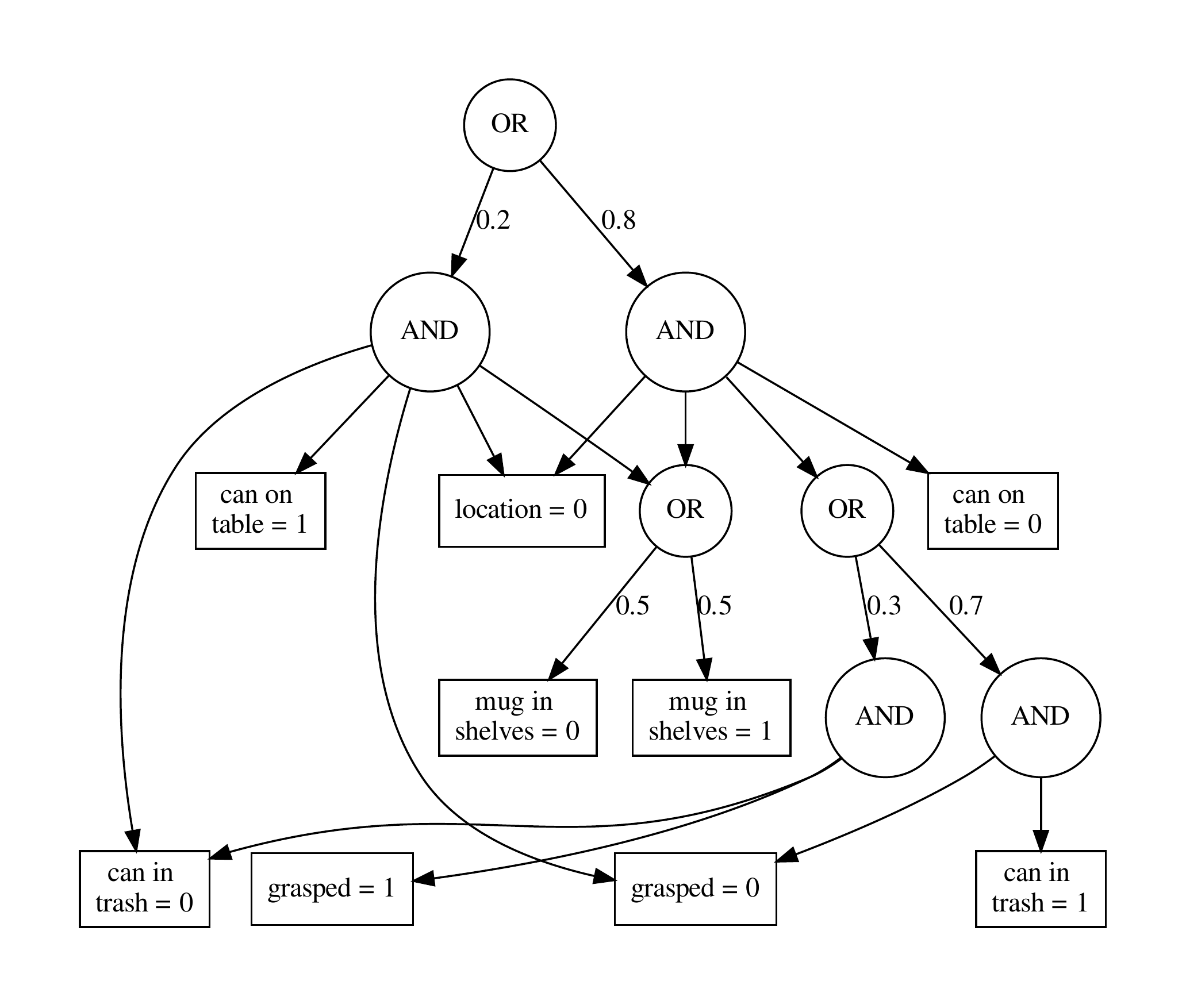}
         \caption{AOBS was normalized. Now \underline{AND}/\nor{} nodes have only \underline{OR}/\nand{} or \nlit{} children, sum of probability factors for each \nor{} node is $1$. }
         \label{fig:aobs6}
    \end{subfigure}%
     \vspace{0em}
    \caption{
    Example of acting on a Belief state. We apply two actions consequently, both with two probabilistic outcomes. The first is "pick can from table" action and the second is "put can into trash" action. We show how belief state is transformed under action application.
      }%
    \label{figs:action}
    \vspace{-1em}
\end{figure*}

     Briefly, the algorithm consists of the following steps. First, from the condition $C$ we select all the literals from AOBS that must be included in the substate ($c=2$ for the example on Figure \ref{fig:aobs1}). Second, we find minimal subgraphs, which contain all variables from action outcomes subspace $\Omega(A)$. By \emph{subminimal subgraphs} we mean such $G(N)$, that corresponding substate has both variables from action and condition and contains at least one physical state on which condition holds true:

	 \begin{equation}
	 \exists s \in S(N): C(s) = true,
	 \label{eq:ms_c}
	 \end{equation}
	 \begin{equation}
	 \Omega(A) \cup \Omega(C) \subseteq \Omega(S(N))
	 \label{eq:ms_vars}
	 \end{equation}
Basically, if a condition holds true on at least one physical state, then the root node of AOBS is a subminimal subgraph. We call subgraph $G(N)$ \emph{minimal}, if it is subminimal and there is no child of $N$ which subgraph is subminimal. If the condition consists of only one single variable relation e.g. $x>0$, then all leaf nodes containing $x$ variable and value above zero are minimal subgraphs.
     As in the case with a single variable condition, there could be multiple such minimal subgraphs. Then, if any of these subgraphs contains physical states, that must not be present in the selection substate, we apply \emph{isolation} procedure. The procedure modifies the selected minimal subgraph $G(N)$ in a way that for all children $M_i$ of $N$ $C(s) = true~\forall s \in S(M_i)$ or $C(s) = false~\forall s \in S(M_i)$ (see Figure \ref{figs:action} for an example). Then, we can modify each isolated subgraph according to the action definition and then apply action effects to isolated substates. Lastly, we normalize graph structure removing redundant \nand{} and \nor{} nodes. As an extra step, greedy optimization could be performed, to reduce the size of the graph (Section \ref{s:greedy}). Let us define a few routines which will form all the steps above.

	 \subsection{Labeling procedure}\label{p:labeling} In this procedure, we want to label each node $N$ with respect to condition $C$. We put a label \lI (\textit{included}) to a node $N$ if all substate $S(N)$ of $\forall C(s) =true, s\in S(N)$, we might later act on the whole substate $S(N)$. Label \lE (\textit{excluded}) corresponds to $\forall C(s) = false, s\in S(N)$. No physical substate from $S(N)$ belongs to $S(C)$ (all the substate should be \emph{excluded}). Otherwise, we label it as \lM (mixed). Nodes with labels \lI and \lM would satisfy first part of subminimal subgraph definition (see eq. \ref{eq:ms_c}). We define \emph{labeling} function as a recursive one, which follows the rules below. \nand{} node should be labeled with \lI iff all its children are labeled with \lIn, with \lE if at least one child is labeled with \lEn,  and labeled with \lM otherwise. \nor{} node is \lI or \lE iff all its children are \lI or \lE respectively. Otherwise, \nor{} node is labeled with \lMn. \nlit{} nodes labels are defined by condition as we can directly evaluate condition function $C$ on \nlit{} substate. In the case that condition does not depend on some variable $v_i \not\in \Omega(C)$ all literals for $v_i$ should have \lI label. Detailed description is given in Alg. \ref{alg:colorize}. Due to limited space, we would not place other routines and functions in the paper, but we have included it in the implementation.

     \subsection{Finding node variable subspaces} To check the second part of subminimal subgraph definition (see eq. \ref{eq:ms_vars}), we need to know $\Omega(S(N))$. It can be found recursively. For \nand{} node $\Omega(S(N))$ is a sum of all children variables, for \nor{} node this set is equal to any of its child. For Literal node, we should include only the literal's variable $N \in Lit(G) \Rightarrow \Omega(N) = \{v(N)\}$.
     \subsection{Finding minimal subgraphs} With the procedures above we can easily find minimal subgraphs. If the node's subset includes at least one physical state $s$ such that, $C(s) = true$, then it was marked with either \lI (included) or \lM (mixed) labels. Having found $\Omega(S(N))$ for each $N$ we can directly check if eq. \ref{eq:ms_vars} is satisfied for $N$ and not satisfied for all its children. Let us show that at least one minimal subgraph exist if condition $C$ holds true on at least one physical state from the whole belief state. Note, that if there $C$ holds true at least at one physical state from the whole belief state, then the root node shall be labeled with either \lI or \lM. For each node $N$  by definition of labeling and finding variable subspaces procedures following statements are satisfied:
	\begin{equation}
	 \forall M \in Children(N), \Omega(S(N)) \supseteq \Omega(S(M)),
	\end{equation}
	 \begin{multline}
	 N \not\in Lit(G), Label(N) = \mathbb{I} \Rightarrow \\ \forall M \in Children(N), Label(M) = \mathbb{I},
	 \end{multline}
	 \begin{multline}
	 N \in And(G), Label(N) = \mathbb{M} \Rightarrow \\ \forall M \in Children(M), Label(M) \in \{\mathbb{I}, \mathbb{M}\},
	 \end{multline}
	 \begin{multline}
	 N \in Or(G), Label(N) = \mathbb{M} \Rightarrow \\ \exists M \in Children(M) , Label(M) \in \{\mathbb{I}, \mathbb{M}\},
	 \end{multline}
	 
From this we can deduce that if we go down from root labeled with \lI or \lM, we always find  at least \lI or \lM child, and as the cardinality of $\Omega(S(N))$ does not increase as we go down from root node, and, at some point, we always find a minimal subgraph. Note, that it is possible that the subgraph we obtained contains not only variables from $\Omega(C) \cup \Omega(A)$, but some extra variables.
\par
Let us show why we act on \emph{minimal} subgraphs. Applying action $A$ on a substate of belief state $B$ selected by condition $C$  could be described as following (here $\overline{S}$ is relative to the whole belief state complement of $S$):
\begin{align*}
\{p_i, s_i \in B | C(s_i) = true \}|_{\Omega(B)/\Omega(A)} \bigtimes A = \\
\bigcup_j \{\overline{S_j} \times S_j | C(\overline{S_j} \times S_j) = true \}|_{\Omega(B)/\Omega(A)} \bigtimes A \overset{\mathrm{\Omega(S_j) \supseteq \Omega(C)}}{=}\\
\bigcup_j \{\overline{S_j}|_{\Omega(\overline{S_j}/\Omega(A)} \times \{S_j|_{\Omega(S_j)/\Omega(A)}\| C(S_j) = true\}\} \bigtimes A =
\end{align*}
\begin{align*} \overset{\mathrm{\Omega(S_j) \supseteq \Omega(A)}}{=} \bigcup_j \{\overline{S_j} \times S_j|_{\Omega(S_j)/\Omega(A)} \| C(S_j) = true\} \bigtimes A = \\
\bigcup_j \{\overline{S_j} \times \left(\left(S_j|_{\Omega(S_j)/\Omega(A)}\|C(S_j) = true\right) \times A\right) \}
\end{align*}

Therefore, we find such $S^*_j$ that for some $S_j \subseteq S^*_j, \Omega(S_J) = \Omega(S^*_j), C(S_j) = true, \Omega(S^*_j) \supseteq (\Omega(A) \cup \Omega(C))$. Then, we \emph{isolate} $S_j$ inside $S^*_j$ to be able to act only at $S_j$.

	 \subsection{Substate isolation} In case that the minimal subgraph $G(N)$  is labeled with \lM, we can not act on the whole $S(N)$. To ensure that we act only on the substate, whether $C$ holds true, we modify the $G(N)$ until we have equivalent subgraph $G(N')$ started from \nor{} node $N'$, and each of its children is labeled with either \lI or \lE label (we call two subgraphs $G(N)$ and $G(N')$ \emph{equivalent} if $S(N) = S(N')$). Since we did this modification, we could act on all the children labeled \lI leaving children with \lE untouched. We can perform such \emph{isolation} recursively in a way that if an internal node has \lM children, we first perform isolation for each of them. If \nor{} node labeled with \lM, but all its children are either \lE or \lI, we do not modify it. If \nor{} node $N$ has \nor{} child $M$ with \lM label, we will simply add children of $M$ to $N$ (multiplying the probability factors of $M$ to factor from $N$ to $M$ edge).  If \nor{} node has a \nand{} child $M$ with \lM label, after isolation procedure, $M$ will be modified to \nor{} node. If \nand{} node is labeled with \lM, we have to replace it with \nor{} node, with two children. The first children shall be an \nand{} node with \nor{} nodes from the original node but cut to only \lI children. It exactly describes the Cartesian product of substates whether $C$ holds true. All the rest substates go to the second children. If there were any \lI children of original \nand{} node, they should be added to both children of the new \nor{} node.
	
	 \begin{figure}[t]
    \centering
    \begin{subfigure}[t]{0.99\columnwidth}
        \centering
\includegraphics[width=0.75\columnwidth]{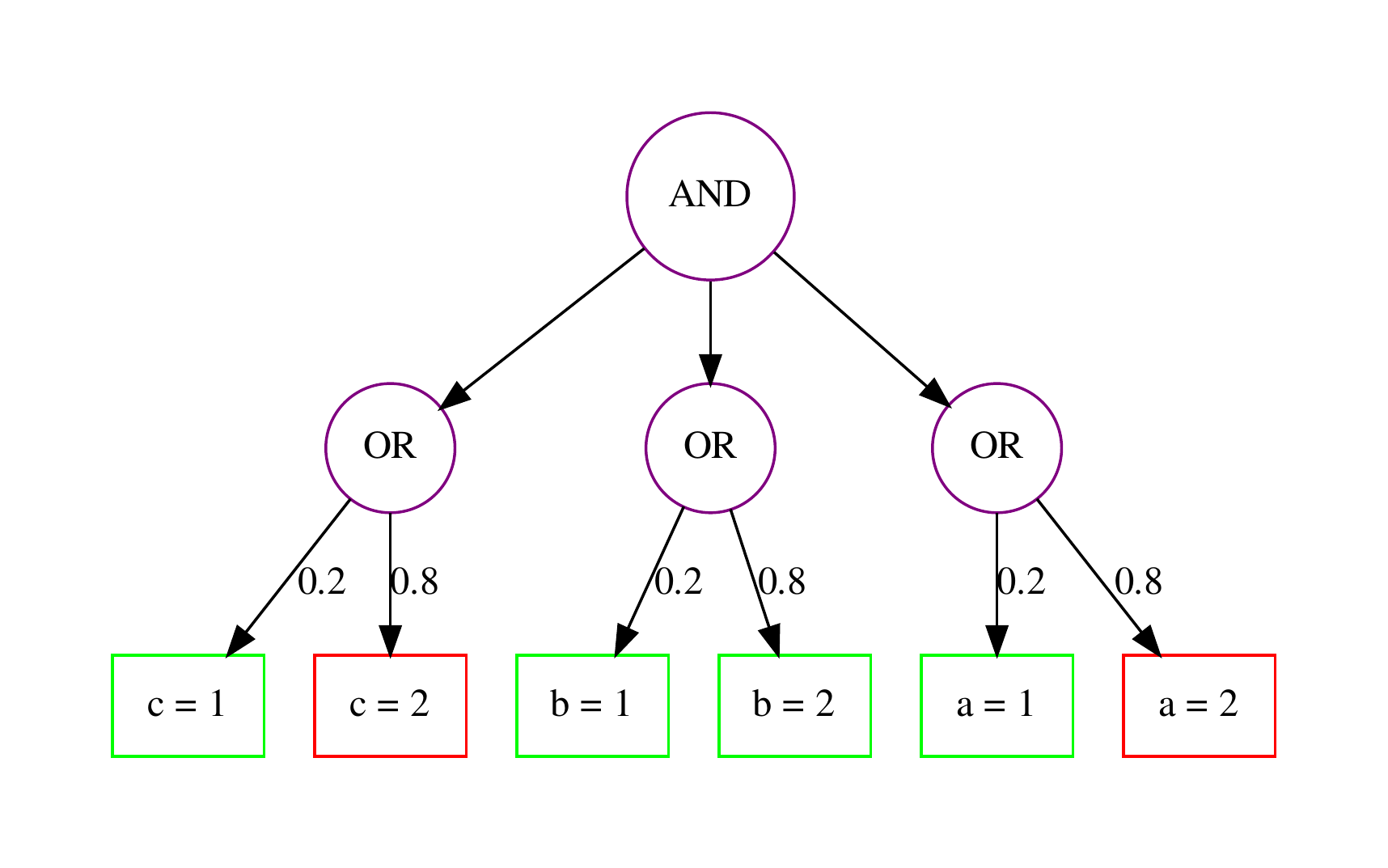}
         \caption{\nand{} node labeled with \lM which should be isolated. \lI painted as green, \lE as red, and \lM as purple. }
         \label{fig:isol_before}
    \end{subfigure}%
    \\
    \begin{subfigure}[t]{0.99\columnwidth}
    \centering
\includegraphics[width=0.75\columnwidth]{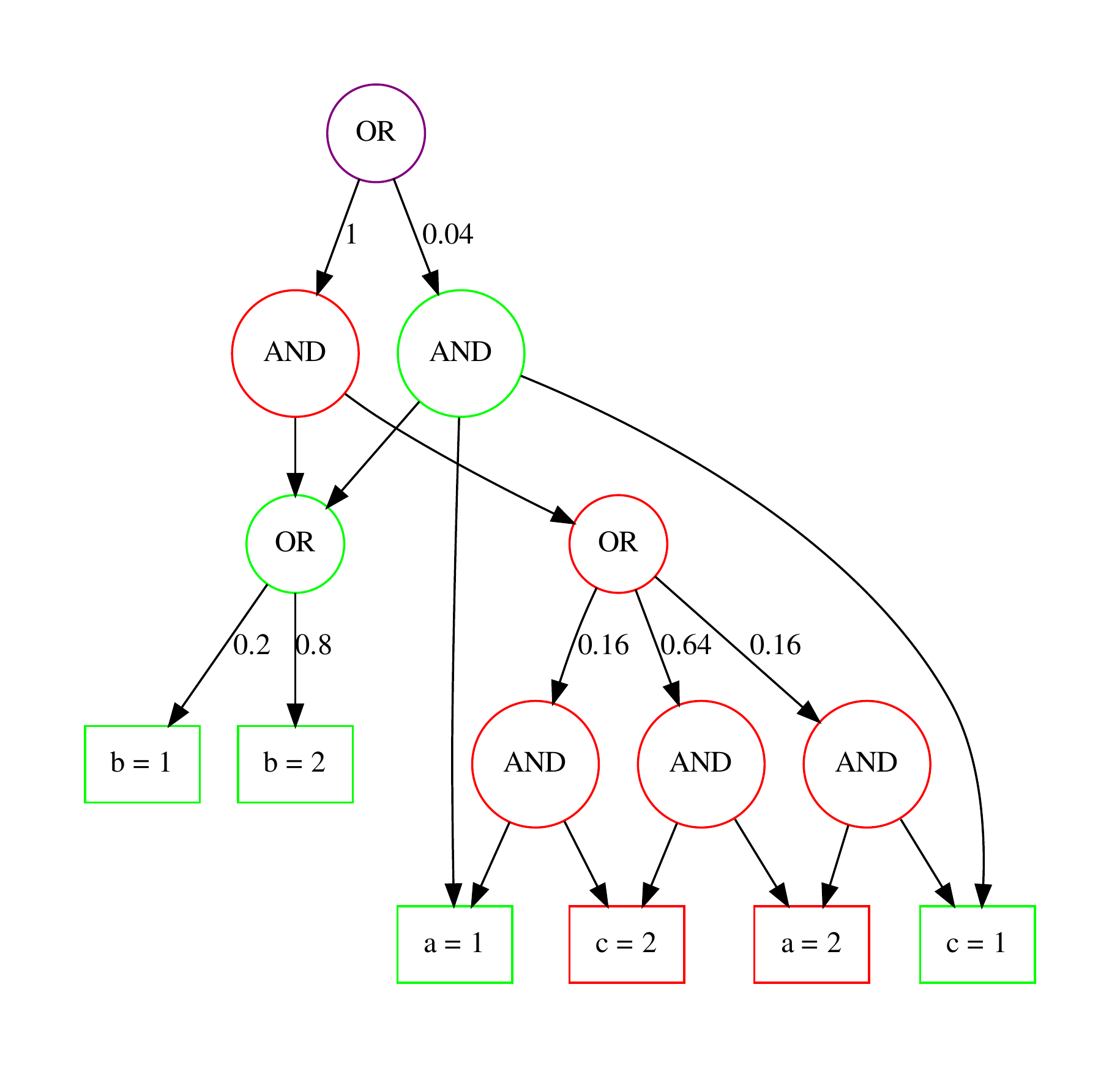}
   \caption{Equivalent to Fig. \ref{fig:isol_before} subgraph. Now, it is rooted in \nor{} node with \lE and \lI children. We should act on the \lI (green) child.}
            \label{fig:isol_after}
    \end{subfigure}
    \vspace{0em}
    \caption{
    	Adding an \nand{} node child for another \nand{} node does not modify structure of statespace and vice versa. This operations could be used for greedy optimization and AOBS normalization after action applied.
      }
      \vspace{-1.5em}
    \label{fig:isol}
\end{figure}
	 \subsection{Removing action variables from isolated subgraphs} Note that we can safely just erase from subgraphs all literals that belong to the action subspace. As action outcomes are state-independent, they will be the same for each physical state from selected substate. Hence that, we reduce the statespace of isolated subgraphs $G(N)$ recursively erasing all $M \in G(N), \Omega(S(M)) \subseteq \Omega(A)$.
     \subsection{Applying action} Now we can add \nand{} node with isolated subgraph and action outcomes as children.
     \subsection{Normalize} When we isolate the subgraph, it could happen that we have \nor{} children of \nor{} nodes (and \nand{} children of \nand{} nodes) after isolation procedure and sums of probabilities for some \nor{} node children edges are not $1$. As we can safely cut a part of \nand{} node to a \nand{} child (same for \nor{}) and vice versa, the normalization procedure becomes trivial. In fact, this property allows performing additional optimization of the graph size (Section \ref{s:greedy}).
     \subsection{Notes on acting procedure}
         We conclude this section with some notes on the procedures we defined above.
     \paragraph{Applying multiple actions simultaneously} Sometimes we must simultaneously apply different actions $A_i$ with for different substates described by conditions $C_i$, whether these substates do not intersect $S(C_i = true) \wedge S(C_j = true) = \emptyset$. As conditions do not intersect, after finding and isolating minimal subgraphs for actions one by one, we would be able to apply actions correctly. The precise procedure for that is out of the scope of this work.
     \paragraph{State dependent actions} We mentioned that described acting procedure is limited to state independent actions. However, we can turn state-dependent action into a set of state independent actions adding dependent variables to conditions. This set of actions must be applied simultaneously.
     \paragraph{More efficient procedures} We described the procedure of finding minimal subgraphs using two recursive functions for the simplicity of explanation and implementation. Even though they do not exceed $O(|G|)$ complexity ($|G|$ - the size of AOBS graph), they could be more efficiently implemented. For example, labeling procedure (Section \ref{p:labeling}) could be done in a non-recursive manner as it is based on a breadth-first search.
	\begin{figure}[t!]
    \centering
    \begin{subfigure}[t]{0.49\columnwidth}
        \centering
\includegraphics[width=0.9\columnwidth]{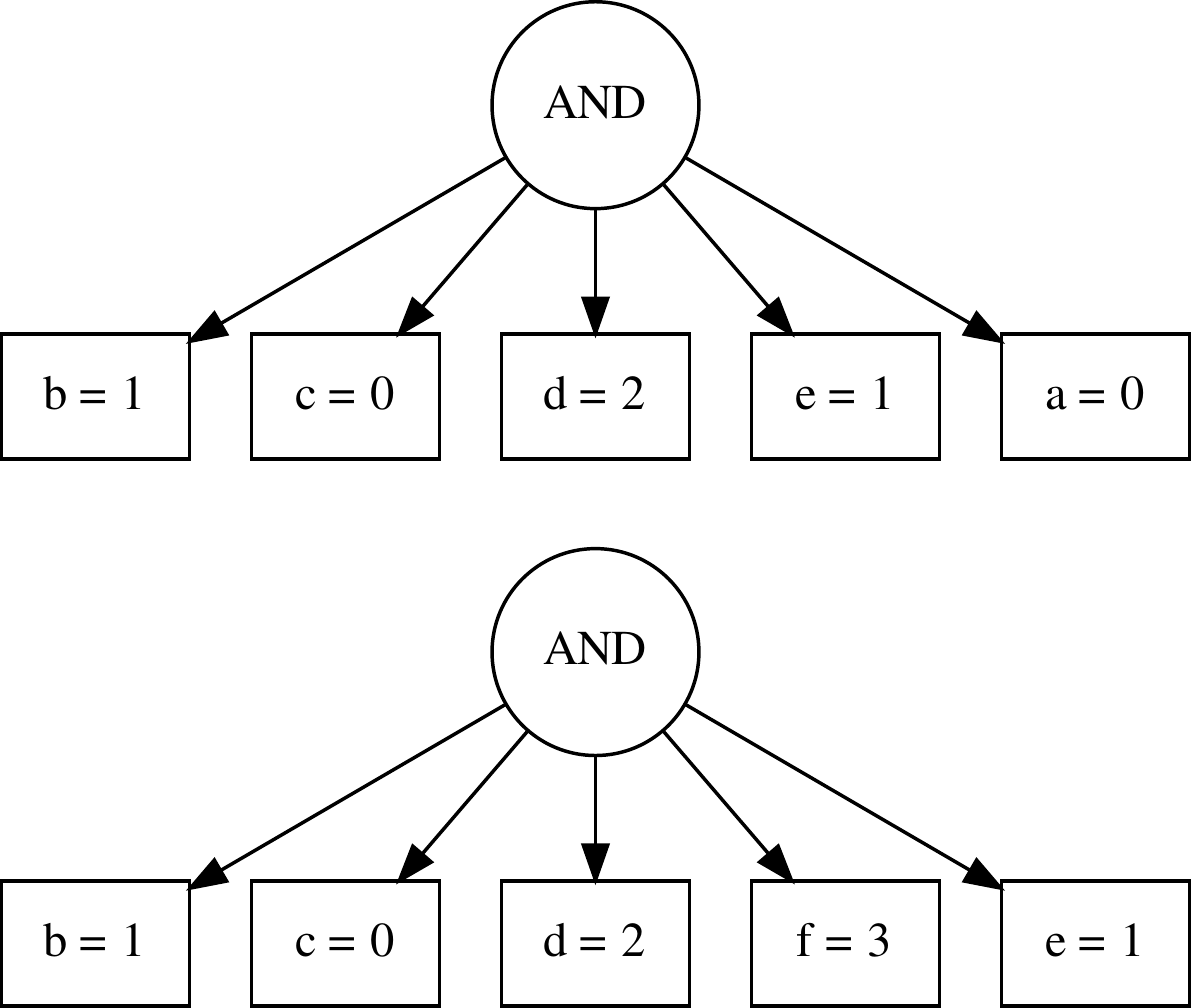}
         \caption{Two \nand{} nodes with non zero children intersection.}
         \label{fig:two_and}
    \end{subfigure}%
    ~
    \begin{subfigure}[t]{0.49\columnwidth}
\includegraphics[width=0.9\columnwidth]{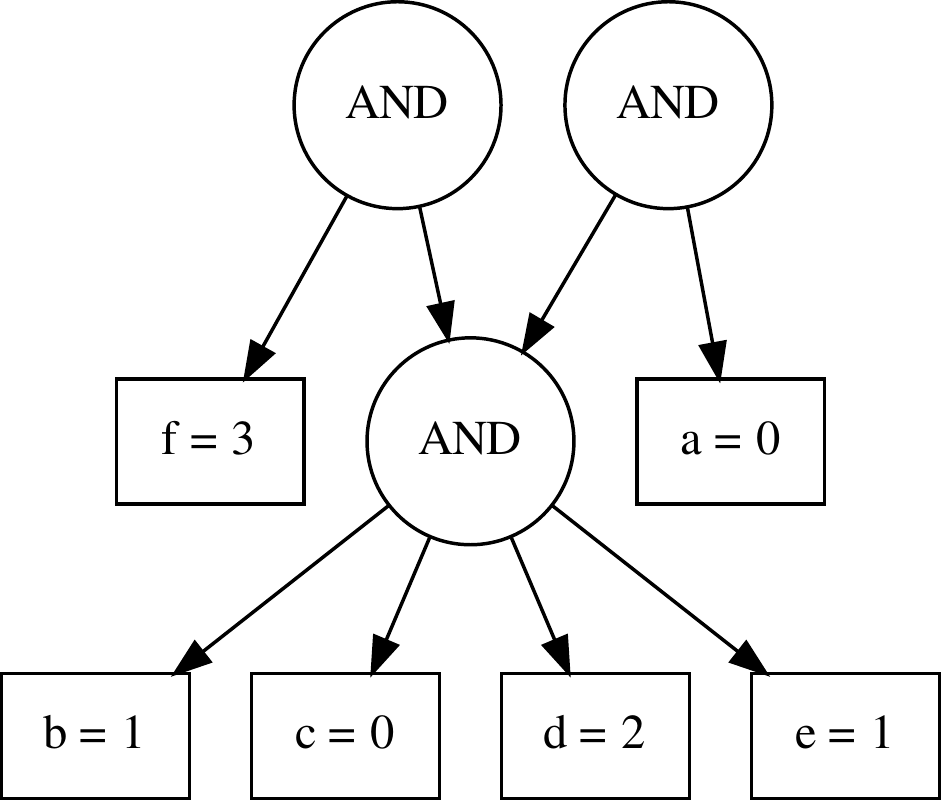}
   \caption{Same two \nand{} nodes (top), but common part of children moved to a new \nand{} node (bottom).}
            \label{fig:merged_and}
    \end{subfigure}
    \vspace{0em}
    \caption{
    	Adding an \nand{} node child for another \nand{} node does not modify structure of statespace and vice versa. This operations could be used for greedy optimization and AOBS normalization after action applied.
      }
    \label{fig:non_unique}
    \vspace{-1.5em}
\end{figure}

\section{Greedy Optimization of And/Or Nodes}
\label{s:greedy}
Sometimes different \nand{} (or \nor{}) nodes have multiple common children.  If we split some \nand{} (or \nor{}) nodes in order to reuse their parts in the other \nand{} nodes of AOBS, we can reduce the total size of the graph. Formally, we have set of sets of elements $W = {S_i, 0 \leq i < N < \infty}$, we want to minimize $\sum_i (|S_i| + C)$ by replacing some elements from $S_i$ to its subset $S_j \subset S_i$ and adding $S_j$ to $W$ if $S_j \not\in W$. $C$ is the cost of having one extra node in memory. This problem belongs to an area of combination optimization. In this work, we applied a greedy approach, in which we simply sorted nodes by their $|S_i|$, looked for the biggest intersection with others, split the nodes and inserted the results back to the queue. If there are no intersections with cardinality higher than the threshold $T$ found, we stop the optimization routine.
\section{Evaluating Conditions on \\ And Or Belief State}
\label{s:conditions}
Another routine operation over a belief state is to calculate a probability of certain conditions holding true. We again limit ourselves to conditions factorized over variables as logical and over single argument functions (Eq. \ref{eq:C}). Probability could be calculated in a way that is similar to the \emph{variablize}  or \emph{labeling} procedure defined in Alg. \ref{alg:colorize}. For each node $N$ we will calculate a probability of $C$ holding true on a substate $S(N)$. For the literal $a=\alpha$, we return $0$ if the corresponding function holds false $f_a(\alpha) = false$ (Eq. \ref{eq:C}) or $1$ otherwise ($f_a(\alpha) = true$ or $a \not\in \Omega(C)$. Then, for the \nand{} node probability is a product of its children probabilities, while for \nor{} node the probability is a sum. Having started this recursive procedure from the root, it will return the probability of $C$ holding true on the whole belief state. A similar task is to select all the physical states $s_i$ (and their probabilities $p(s_i)$) by the condition and return them as a plain collection. It can be done again recursively, now we \nor{} nodes should sum up collections of substates, while \nand{} node should make a Cartesian product (just by their definition).

\section{Experimental Evaluation}
We implemented the described concept of AOBS as a Python package available online.\footnote{https://github.com/safoex/bsagr} It includes code for acting on a probabilistic belief state (Section \ref{s:acting}), evaluating conditions (Section \ref{s:conditions}), and few other helpful procedures. We provide a visualization of the AOBS based on a \emph{graphviz} dot language \cite{Ellson03graphvizand}. Nodes storage and merging isomorphic subgraphs is handled by hashing. \nand{} nodes are lists of hashes $h_i$, each hash uniquely defines some other node from the AOBS. Similarly, \nor{} nodes are lists of tuples $(p_i, h_i)$, where $p_i$ is a probability of a substate defined by node hashed with $h_i$. Hence, if two nodes in the graph have the same isomorphic subgraph, they will be merged automatically having the same hash $h_i$.
\subsection{Correctness check} In order to check the correctness of developed algorithms, we implemented a simple probabilistic belief state representation in tabular form. Then, we generated a series of random physical states that initiated belief states in both forms (AOBS and tabular) and applied random exploration with the same sequence of actions to each of the representations. Then, we recovered tabular  representation from AOBS and compared it to original tabular form. In all cases, results were the same up to neglectable differences in $p_i$ probability values due to the numerical stability of mathematical operations in real computations.
\subsection{Numerical evaluation}
    As the goal of developed AOBS was to keep the size of belief state representation as small as possible, our natural benchmark is the graph size compared to other possible representations. The graph size is counted by $N_{AOBS} = |G_{AOBS}| = |E| + N_{And} + N_{Or} + 2 \cdot N_{Lit}$. Naively belief state could be implemented as a plain collection of tuples (probability, physical state). In this case, number of naive states is product of number of states in a belief states and variable statespace cardinality: $N_{naive} = |V| \cdot N_{states}$. We compare it against representation of the belief state (without probabilities) in a form of BDDs. We do not compare the memory footprints of the developed program for comparison to be independent of the implementation details.
    \par We used a simulated policy exploration procedure to generate belief states. In each experiment, statespace was formed by $|V|$ discrete variables, each could hold $|U|$ integer values. For the corresponding BDD representation, statespace consisted of $|V|\cdot|U|$ boolean variables, where each boolean variable $v_u$ is true when $v = u, v \in V, 0 \leq u < |U|$ and false otherwise. Each policy exploration started from the randomly generated physical state. Then, we applied a sequence of $N_{actions}$. Each action had $N_{eff}$ outcomes, changing values of $N_{assign}$ variables. To select a substate we used $N_{conditions}$ randomly generated conditions. In all studied cases, AOBS is significantly more efficient than the naive representation. Note, increasing $|U|$ and keeping other parameters the same leads to the higher belief state compression efficiency.

\begin{figure}[t!]
    \centering
    \begin{subfigure}[t]{0.49\columnwidth}
        \centering
\includegraphics[width=0.99\columnwidth]{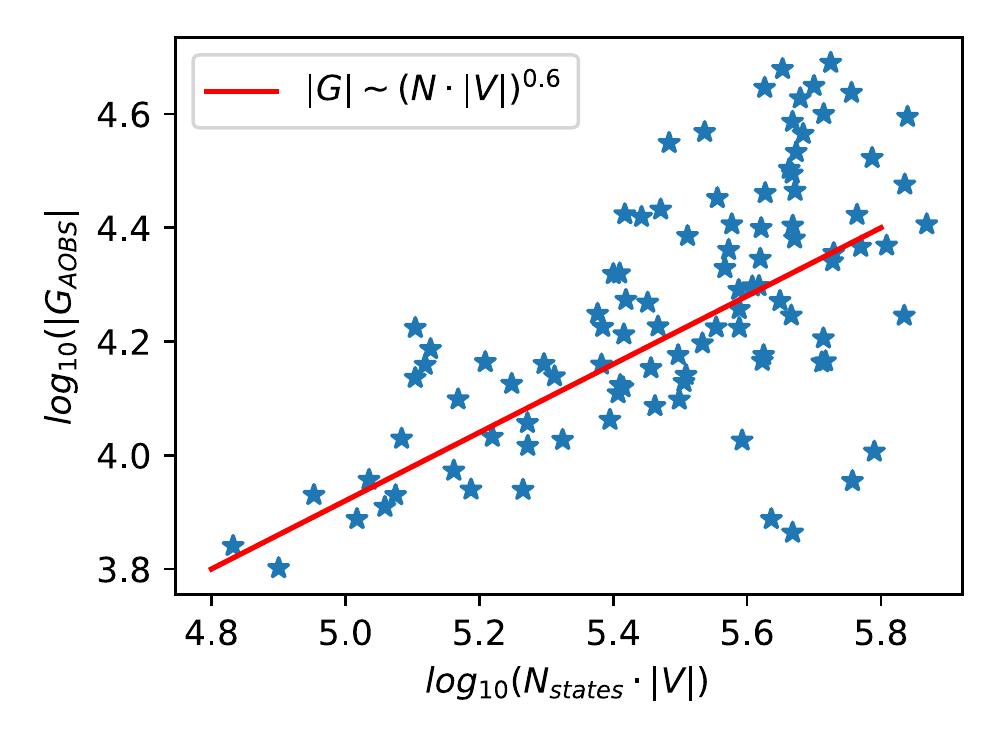}
         \caption{$|G| \sim (N\cdot|V|)^{0.6}$ for $|U| = 2$}
         \label{fig:exp_n06}
    \end{subfigure}%
    ~
    \begin{subfigure}[t]{0.49\columnwidth}
\includegraphics[width=0.99\columnwidth]{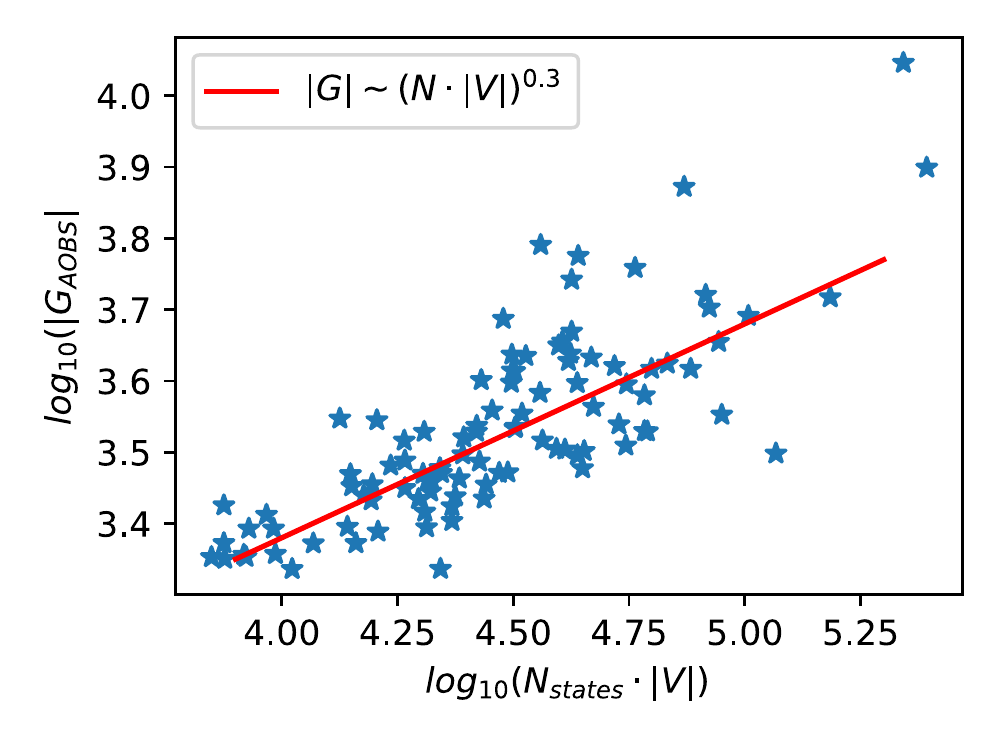}
   \caption{$|G| \sim (N \cdot |V|)^{0.3}$ for $|U| = 8$}
            \label{fig:exp_n03}
    \end{subfigure}
    \vspace{0em}
    \caption{
    Comparison between naive belief state and AOBS sizes after simulated statespace exploration. Increasing $|U|$ leads to more sparse belief state. For such belief states, AOBS scales better. Plots are in logarithmic scale. $|V| = 30$, $N_{eff} = 3$, $N_{assign} = 3$, $N_{actions} = 35$, $N_{conditions} = 3$
      }
    \label{fig:plots}
\end{figure}

\begin{figure}[t!]
    \centering
    \begin{subfigure}[t]{0.49\columnwidth}
        \centering
\includegraphics[width=0.99\columnwidth]{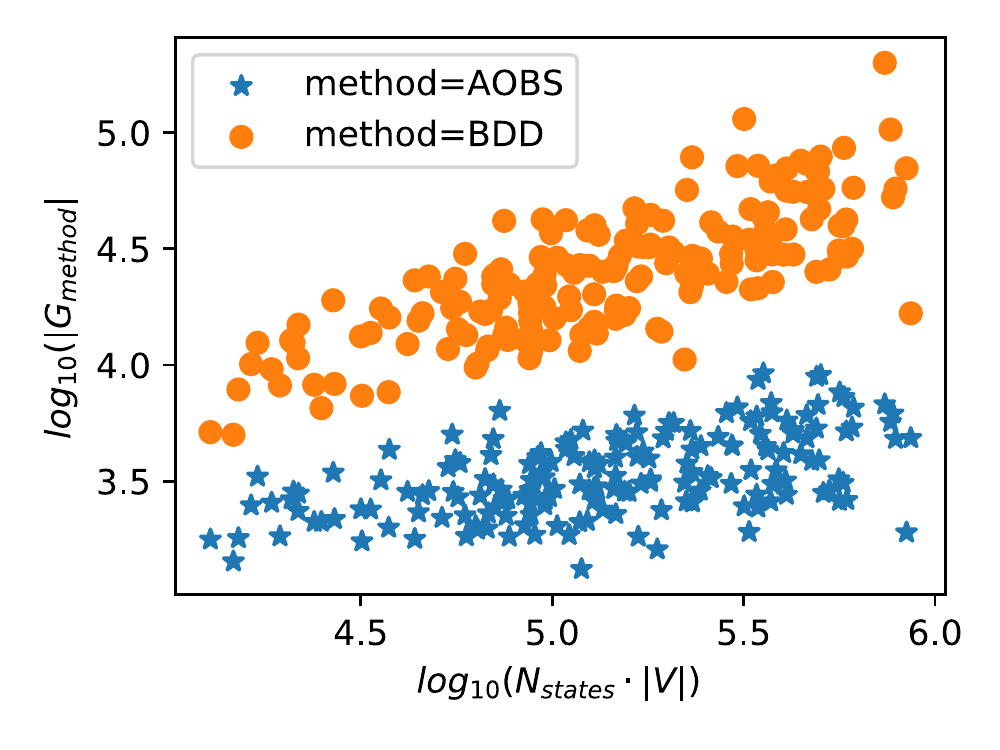}
         \caption{$|V| = 50$. A distribution of graph sizes for 200 simulated random statespace explorations, logarithmic scale. AOBS outperforms BDD in terms of graph size. $|U| = 4, N_{eff} = N_{assign} = N_{conditions} = 3, N_{actions} = 20$  }
         \label{fig:eval50}
    \end{subfigure}%
    ~
    \begin{subfigure}[t]{0.49\columnwidth}
\includegraphics[width=0.99\columnwidth]{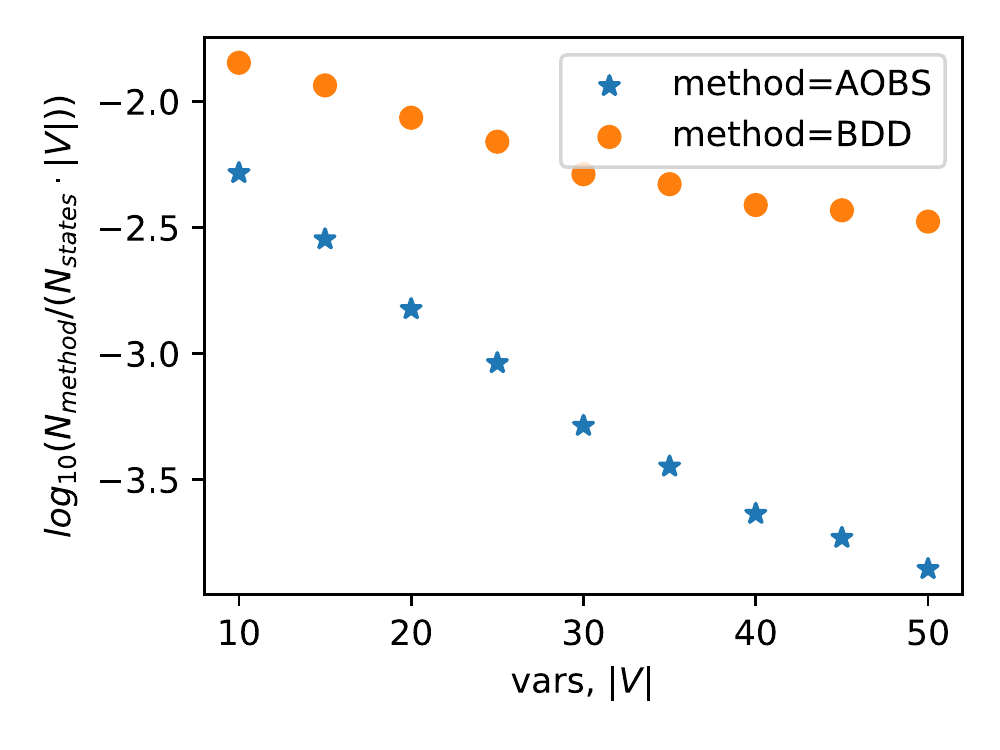}
   \caption{Averaged compression ratio $(N_{states}|V|)/|G|_{method}$ for AOBS and BDD for different statespace sizes $15 \leq |V| \leq 50$. AOBS scales better for larger models statespace.}
            \label{fig:eval_vars}
    \end{subfigure}
    \vspace{0em}
    \caption{
    Comparison between naive belief state, BDD, and AOBS sizes after simulated statespace exploration. With larger number of  variables in the belief state, AOBS performs much better then BDD.
      }
    \label{fig:plots_V}
\end{figure}

\begin{figure}[t!]
    \centering
    \begin{subfigure}[t]{0.49\columnwidth}
        \centering
\includegraphics[width=0.99\columnwidth]{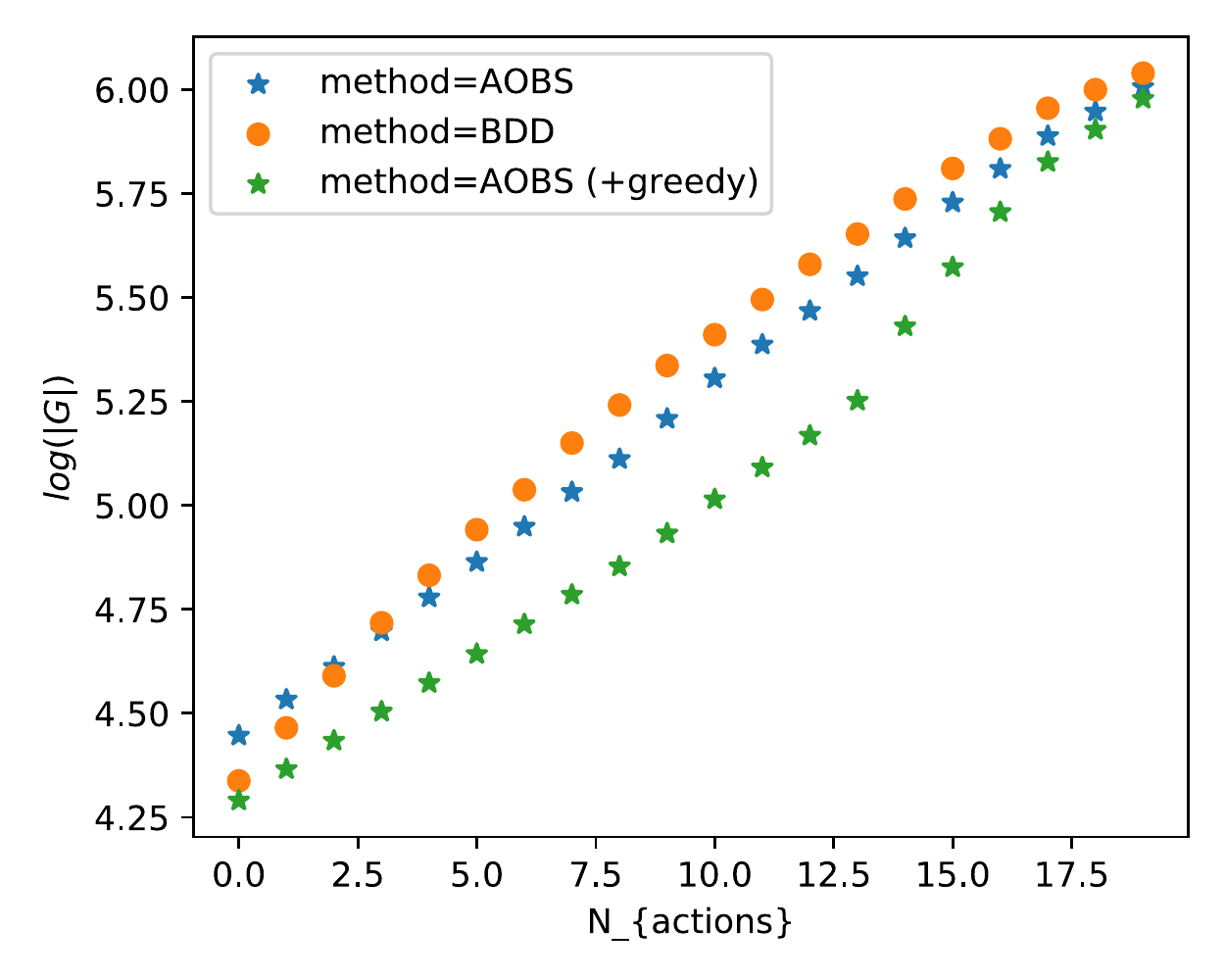}
         \caption{$|U| = 2$. Due to limited value space size, BDD performs quite well thanks to elimination rule.   }
         \label{fig:eval_U2}
    \end{subfigure}%
    ~
    \begin{subfigure}[t]{0.49\columnwidth}
\includegraphics[width=0.99\columnwidth]{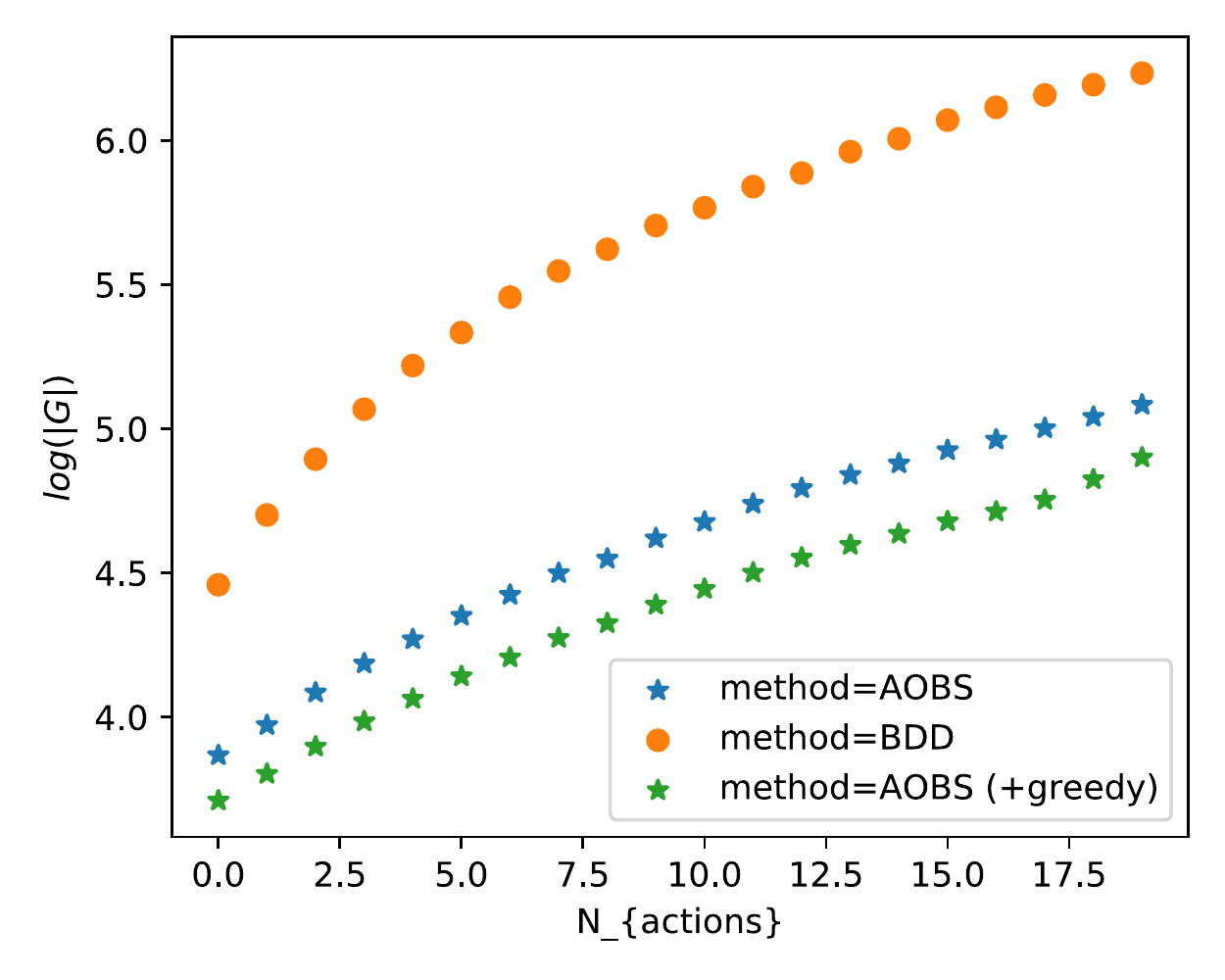}
   \caption{$|U| = 8$. When BDD has less opportunities for elimination, AOBS wins by order of magnitude thanks to order independence.}
            \label{fig:eval_U8}
    \end{subfigure}
    \vspace{0em}
    \caption{
    Comparison between BDD and AOBS graph sizes average for different number of actions applied $N_{actions}$. Other parameters $|V| = 40, N_{eff} = N_{assign} = N_{conditions} = 3$
      }
    \label{fig:plots_N}
    \vspace{-1.5em}
\end{figure}
\paragraph{BDD vs AOBS} We compared the sizes of the AOBS graph with BDD graph sizes for the same experiments. BDD are capable of representing a belief state without probabilities of physical states $p(s)$  as a boolean function $b(s) := p(s) > 0$.  As variables in our simulation are not boolean, we follow a standard approach for Multivalued Decision Diagrams\cite{129849}, encoding them to BDDs. We used the available package to evaluate BDD performance\footnote{https://github.com/tulip-control/dd}. Variable assignment $v_i = u_j$ corresponds to $\forall k \neq j, {v_i}_{k} = false; {v_i}_{j} = true$. An action outcome consists of several conjugated variable assignment $\bigwedge_k v^k_i =u^k_j$, and an action consists of several action outcomes $A = \bigvee_l \bigwedge_k v^{lk}_i = u^{lk}_j$. If condition is described by boolean function $C$, belief state by function $b$, then the belief substate where $C = true$ is $b \bigwedge C$. The result of applying an action is $(b \bigwedge \neg C) \bigvee \left((b \bigwedge C)|_{\Omega(A)} \bigwedge A\right)$.
\par We want to highlight here important evaluation results. We showed that AOBS nonlinearly compresses belief state representation (see Figure \ref{fig:plots}). For most of the parameter combinations, AOBS graph size was lower even than BDD graph size with elimination rule enabled (see Figure \ref{fig:plots_V}). We would like to point out again, that BDD could not be straightforwardly applied in the probabilistic belief state case (see \ref{s:rel-works}) and serves here just as baseline. For the sparse statespace AOBS size was more than $1000$ times compared to naive belief state representation in some experiments (Figure \ref{fig:eval_vars}). What is more important, AOBS was less sensitive to parameter scaling (for statespace scaling at Figure \ref{fig:plots_N}). This proves that AOBS belief state representation could be competitive in the case of belief state without probabilities too.

\section{Conclusion}
We developed a novel probabilistic belief state representation based on an And Or direct acyclic graph named AOBS. We showed how to apply actions on a belief substate in AOBS form and calculate the probability of a given condition. We showed that the size of the AOBS graph is much smaller than the size of the collection of a physical state in a simulated random statespace exploration experiment. We compared the size of AOBS to the size of a BDD representation of belief states without probabilities. Results reveal that AOBS scales better for bigger models outperforming BDDs, and therefore could be applied for nondeterministic belief state as well. 

    \bibliography{root}

    \bibliographystyle{ieeetr}
\end{document}